\documentclass[10pt,twocolumn,letterpaper]{article}

\usepackage{cvpr}              

\definecolor{cvprblue}{rgb}{0.21,0.49,0.74}
\usepackage[pagebackref,breaklinks,colorlinks,allcolors=cvprblue]{hyperref}
\usepackage{algorithm}%
\usepackage{algorithmicx}%
\usepackage{algpseudocode}%
\usepackage{colortbl}
\definecolor{secondcolor}{RGB}{220,230,240}
\definecolor{firstcolor}{RGB}{241,220,219}

\graphicspath{{Figure/}}
\def\paperID{****} 
\def\confName{CVPR}
\def\confYear{2024}

\title{ReFusion: Learning Image Fusion from Reconstruction \\ with Learnable Loss via Meta-Learning}

\author{
        Haowen Bai$^{1}$\quad
        Zixiang Zhao$^{1,2}$\footnotemark[1]\quad
        Jiangshe Zhang$^{1}$\footnotemark[1]\quad
        Yichen Wu$^{3}$\\
        Lilun Deng$^{1}$\quad
        Yukun Cui$^{1}$\quad
        Baisong Jiang$^{1}$\quad
        Shuang Xu$^{4}$\quad\\[1mm]
        $^{1}$Xi’an Jiaotong University\quad
        $^{2}$ETH Z\"urich\\
        $^{3}$City University of Hong Kong\quad
        $^{4}$Northwestern Polytechnical University\\
        {\tt\small hwbaii@stu.xjtu.edu.cn}
}

\begin{document}
\maketitle

\footnotetext[1]{Corresponding authors.\raisebox{0.5ex}{\textasteriskcentered}}

\begin{abstract}
Image fusion aims to combine information from multiple source images into a single one with more comprehensive informational content.
Deep learning-based image fusion algorithms face significant challenges, including the lack of a definitive ground truth and the corresponding distance measurement. Additionally, current manually defined loss functions limit the model's flexibility and generalizability for various fusion tasks.
To address these limitations, we propose \textbf{ReFusion}, a unified meta-learning based image fusion framework that dynamically optimizes the fusion loss for various tasks through source image reconstruction.
Compared to existing methods, ReFusion employs a parameterized loss function, that allows the training framework to be dynamically adapted according to the specific fusion scenario and task.
ReFusion consists of three key components: a fusion module, a source reconstruction module, and a loss proposal module.
We employ a meta-learning strategy to train the loss proposal module using the reconstruction loss. 
This strategy forces the fused image to be more conducive to reconstruct source images, allowing the loss proposal module to generate a adaptive fusion loss that preserves the optimal information from the source images.
The update of the fusion module relies on the learnable fusion loss proposed by the loss proposal module.
The three modules update alternately, enhancing each other to optimize the fusion loss for different tasks and consistently achieve satisfactory results.
Extensive experiments demonstrate that ReFusion is capable of adapting to various tasks, including infrared-visible, medical, multi-focus, and multi-exposure image fusion.
The code is available at~\url{https://github.com/HaowenBai/ReFusion}.
\end{abstract}

\section{Introduction}
\label{sec1}
Image fusion plays a crucial role in computer vision and image processing by merging multiple views of a scene into a unified image that conveys enhanced, multi-dimensional information.
The fusion process leverages the strengths of each source image while mitigating the weaknesses inherent in uni-modal imaging or specific imaging techniques, thereby improving the overall quality and utility of the fused image.
The core advantage of image fusion lies in its capacity to synthesize information from disparate imaging sources that may differ in sensor type, focus area, or exposure time. 
Its wide variety of applications span medical imaging~\citep{DBLP:journals/inffus/JamesD14,li2023gesenet,litjens2017survey}, remote sensing~\citep{DBLP:conf/cvpr/BandaraP22,DBLP:conf/cvpr/Xu0ZSL021,9913829,10144690}, surveillance~\citep{DBLP:journals/inffus/LiZHWC20}, autonomous vehicles~\citep{DBLP:journals/tcsv/MaikCSP07}, and more~\citep{hu2024cross,zhou2023underwater}.
Specifically, infrared-visible image fusion (IVIF)~\citep{liu2024coconet,wang2024general,li2023deep} preserves the structural thermal radiation details from infrared images and incorporates textural information from visible images.
Medical image fusion (MIF) \citep{xu2021emfusion} integrates information from varying modalities, such as MRI and CT scans, offering a more detailed anatomical perspective.
Multi-focus image fusion (MFIF)~\citep{zhang2021deep,hu2023zmff,liu2022multi,wang2022self} synthesizes fully focused images from pairs of images with distinct focus areas. Similarly, multi-exposure image fusion (MEIF)~\citep{jiang2023meflut,zhang2021benchmarking,DBLP:journals/tip/MaDZFW20,DBLP:journals/tip/MaLYWMZ17} crafts high dynamic range images from a series of low dynamic range images taken at varying exposure levels~\citep{yan2022dual,liu2021benchmarking,zhang2021beyond,xu2024cretinex}.

\begin{figure*}[t]

\centering
\includegraphics[width=\linewidth]{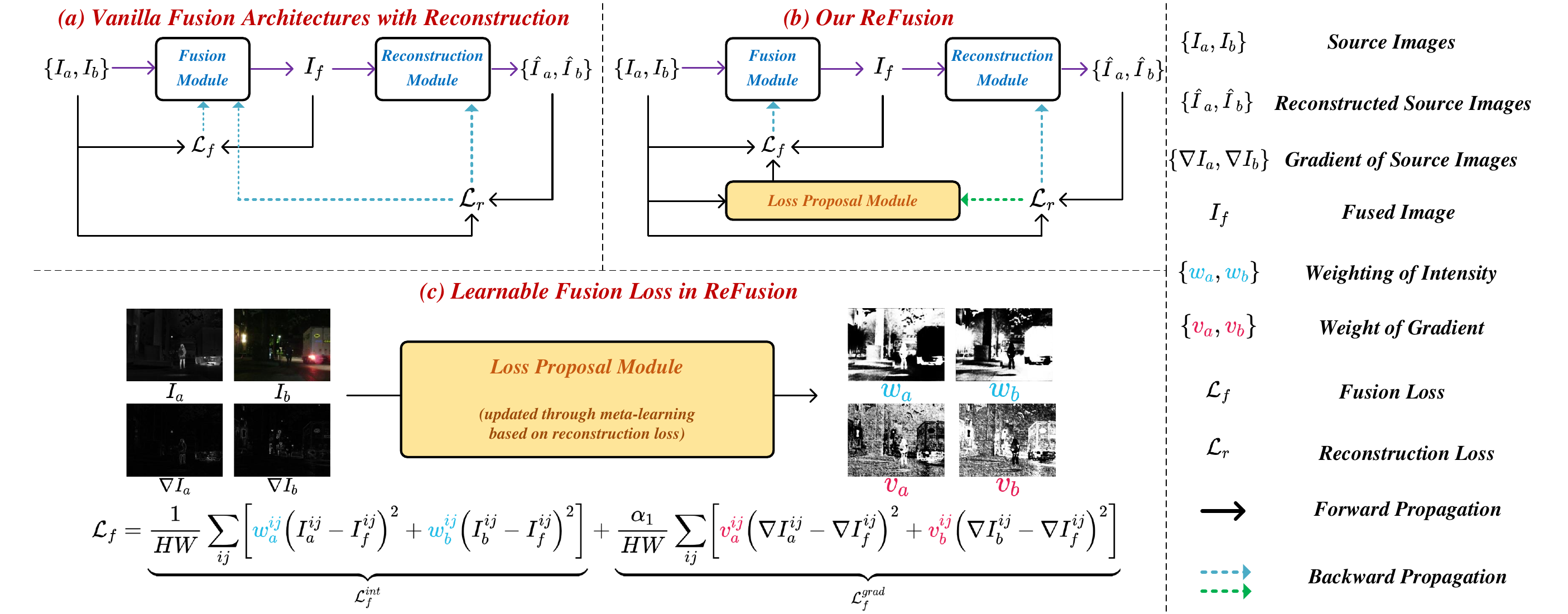}
	\caption{Comparison of the proposed ReFusion framework 
with existing methods, such as ~\citep{DBLP:journals/ijcv/ZhangM21}. \textbf{(a)} In previous fusion methods, reconstruction loss supports the fusion loss during the training of the fusion module. \textbf{(b)} In ReFusion, reconstruction loss is employed to supervise the fusion loss proposal module in a meta-learning manner. This supervision ensures that the fused images more effectively reconstruct the source images, enabling the loss proposal module to propose a fusion loss function that better preserves the source image information. {\textbf{(c)} The parameters of the learnable fusion loss, output by the loss proposal module, where $w_{a}^{ij}+w_{b}^{ij}=1$ and 
$v_{a}^{ij}+v_{b}^{ij}=1$. The fusion loss is learned by supervising the loss proposal module, which dynamically adjusts the parameters based on the specific requirements of the fusion task.}}
\label{RL}
\vspace{-1em}
\end{figure*}

Establishing a unified fusion framework requires a versatile loss function that adapts to the specifics of each task and scenario.
However, the lack of ground truth in image fusion necessitates the devising of manually specified loss functions. 
Predominant practices include using an equal-weighted approximation of the source images within the fused output~\citep{zhao2020didfuse}, or formulating weighting rules tailored to specific tasks~\citep{DBLP:journals/ijcv/ZhangM21} or scenes~\citep{DBLP:journals/inffus/TangYZJM22}.
Additionally, techniques such as adversarial learning have been applied to force the distribution of the fused image closely approximate that of the source images~\citep{ma2019fusiongan,ma2020ddcgan}. 
However, using equal weights or manually determined weights in the objective function is inappropriate due to the varying imaging features and representational information among source images in different fusion tasks.
The variability among different fusion tasks means that existing dynamic weighting strategies~\citep{zhou2021semantic,9151265} cannot be universally applied to all tasks and scenarios.
For instance, the method of dynamically adjusting the source image weights based on scene illumination in IVIF~\citep{DBLP:journals/inffus/TangYZJM22} is ineffective in solving MFIF where illumination conditions of the source images are essentially consistent.
Similarly, the commonly used approach in MFIF, which controls loss weights based on standard deviation~\citep{DBLP:journals/sensors/YanGQM20} or gradient, fails to be effective in tasks that emphasize differences in color, texture, or other information types.
Moreover, the simplistic criterion of selecting maximal values from the source images~\citep{zhao2023cddfuse,tang2022image} risks the omitting of critical information~\citep{tang2023rethinking}. 
Thus, in this paper, we propose a learnable loss function coupled with a learning framework aiming to suit arbitrary fusion tasks. The mechanism is demonstrated in Fig.~\ref{RL}. 
This learnable loss function assigns pixel-wise preferences to the source images based on intensity and gradient aspects, modeled by a neural network and dynamically adjusted during the training process. This design ensures the similarity of fused images and provides a learning space to adapt the fusion loss to various tasks.

{The neural network used for modeling the fusion loss cannot be trained through traditional end-to-end methods, as its output serves to train the fusion module within the fusion loss to assist in generating the fused image. Therefore, in this paper, meta-learning is utilized to develop a learnable fusion loss function that enhances learning algorithms by imparting the skill of adaptive learning, diverging from the static learning trajectory of conventional algorithms~\citep{DBLP:journals/pami/HospedalesAMS22}.}
This adaptive learning approach addresses enduring challenges in deep learning such as data limitations, computational overhead, and the need for improved generalizability.
The focal point of optimization in meta-learning typically revolves around parameters~\citep{finn2017model}, optimization~\citep{li2017meta}, and network architectures~\citep{liu2018darts}, among others.
Within the scope of this paper, we present a learnable loss function formulated through a module named the Loss Proposal Module (LPM). 
We employ the Model-Agnostic Meta-Learning (MAML) approach~\citep{finn2017model}, a prominent meta-learning paradigm, to train the LPM. The trained LPM can then propose an optimal fusion loss to produce well-fused images.

The fundamental purpose of image fusion is to preserve the information from the source images, which also motivates the design of loss functions in much of the previous research~\citep{zhao2023cddfuse,zhao2020didfuse}.
Different from existing methods, we employ the reconstruction loss of the source images to measure the loss of information, which serves as more consistent information preservation criterion for all tasks. 
The underlying insight is that for the source images to be successfully reconstructed, their information must be maximally preserved in the fused image.
This idea has also been used to assist in training fusion networks~\citep{DBLP:journals/ijcv/ZhangM21,DBLP:journals/corr/abs-2211-04877}.
An ideal fusion loss function should inherently guide the network toward producing a fused image that not only integrates the fundamental information of the source images but also simplifies the task of reconstructing the source images.
Therefore, we employ the reconstruction loss of the source images to train the loss proposal module and to obtain the desired fusion loss.
The role of reconstruction loss in our fusion framework is demonstrated in Fig.~\ref{RL}.
In light of these considerations, we introduce an innovative image fusion framework called \textbf{ReFusion}, which provides a learning paradigm designed to achieve the optimal fusion loss for various tasks based on the reconstruction of source images.
This framework is structured around three core components: \emph{1)} an efficient module for effectuating the fusion (the \textit{lightweight fusion module}); \emph{2)} a two-branch module dedicated to the task of reconstruction (the \textit{source reconstruction module}); \emph{3)} a module that proposes loss functions tailored to the specific scenario and task (the adaptive \textit{fusion loss proposal module}).

{The ReFusion learning paradigm consists a cycle of three learning stages, as illustrated in Fig.~\ref{fig:Refusion}. 
The inner update, represented in red~\normalsize{\textcircled{\footnotesize{1}}}\normalsize, and the outer update in green~\normalsize{\textcircled{\footnotesize{2}}}\normalsize, collectively facilitate the update of loss proposal module. The inner update employs the currently proposed fusion loss, and the outer update assesses the effectiveness of fusion loss while optimizing the loss proposal module using the reconstruction loss. This process does not alter the parameters of the fusion and reconstruction modules. The fusion and reconstruction update represented in blue~\normalsize{\textcircled{\footnotesize{3}}}\normalsize~optimizes fusion module and reconstruction module by the currently proposed fusion loss and the reconstruction loss.}

Our main contributions are summarized as follows:

(1) We develop a unified image fusion approach, called ReFusion, which leverages a loss function learned through meta-learning. 
The loss proposal module dynamically proposes suitable loss functions based on the specific scene and task.
Notably, the learning process, situated within a meta-learning framework, relies solely on reconstructing source images, thereby eliminating the need for the ground truth of the fused image.

(2) The outstanding performance of ReFusion is secured through a learnable fusion loss function, which reduces reliance on traditional, complex network architectures. We employ a novel lightweight fusion network architecture that efficiently merges features using interactive feature extraction and gating mechanisms, thus guaranteeing the quality of the fused images with reduced space occupation and computational complexity.

(3) ReFusion consistently delivers high-quality results across diverse image fusion tasks, including infrared-visible, medical, multi-focus, and multi-exposure image fusion. This extensive applicability fully demonstrates the effectiveness and adaptability of our approach in handling various image fusion tasks.

\section{Related Work}\label{Related Work}
\subsection{Deep Learning-based Image Fusion}\label{Deep Learning-based Image Fusion}
Deep learning-based image fusion methods, relying on the powerful feature extraction capabilities of neural networks, have sparked a revolution in the field of image fusion. 
Generally, these methods can be categorized into pixel-level fusion, feature-level fusion, and decision-level fusion.
Pixel-level fusion directly integrates the information of source images, usually resulting in improved image clarity and spatial resolution.
This category also includes GAN-based approach~\citep{ma2019fusiongan,DBLP:conf/cvpr/LiuFHWLZL22} and diffusion-based approach~\citep{zhao2023ddfm}.
In feature-level fusion, features extracted from the source images are combined using specific strategies to synthesize a fused image.
This type of method mainly appears in autoencoder (AE)-based methods~\citep{zhao2023cddfuse,li2021rfn,zhao2020didfuse,li2018densefuse}, which can be considered as a deep learning adaptation of traditional multiscale transform methods~\citep{BULANON200912,JIN20181, DBLP:journals/mta/EKR19,DBLP:journals/sigpro/LiuJWSD14,DBLP:journals/ijon/LiuMD17}.
Decision-level fusion methods directly predict decisions or classifications from the source images, combining them into a fused image, commonly used in multi-focus image fusion.
Some works are not merely limited to the fused image generation, the combination of image fusion with image registration techniques addresses the issue of misalignment in the source images~\citep{xu2022rfnet,huang2022reconet}.
The self-supervised learning paradigm enables the training of fusion networks under conditions lacking paired images~\citep{Liang2022ECCV}.
Combining image fusion with high-level visual tasks such as semantic segmentation~\citep{DBLP:journals/ieeejas/TangDMHM22, tang2022image, Liu_2023_ICCV} and object detection~\citep{DBLP:conf/cvpr/LiuFHWLZL22,zhao2023metafusion} enhance the adaptability of the fused images for downstream tasks.
Unified models~\citep{9151265,Liang2022ECCV,DBLP:journals/ijcv/ZhangM21} explore the essence of fusion problems and propose a universal solution applicable to all fusion tasks.
In this paper, we address the challenges of lacking ground truth and definitive distance measurement in image fusion. Instead of employing an artificially specified loss function, we design the loss proposal module to establish an appropriate fusion loss. This module aims to maximize the retention of information from the source images and is updated by reconstruction loss through meta-learning, thereby enhancing the efficacy of the fusion process.

\begin{figure*}[t]	
	\centering
	\includegraphics[width=\linewidth]{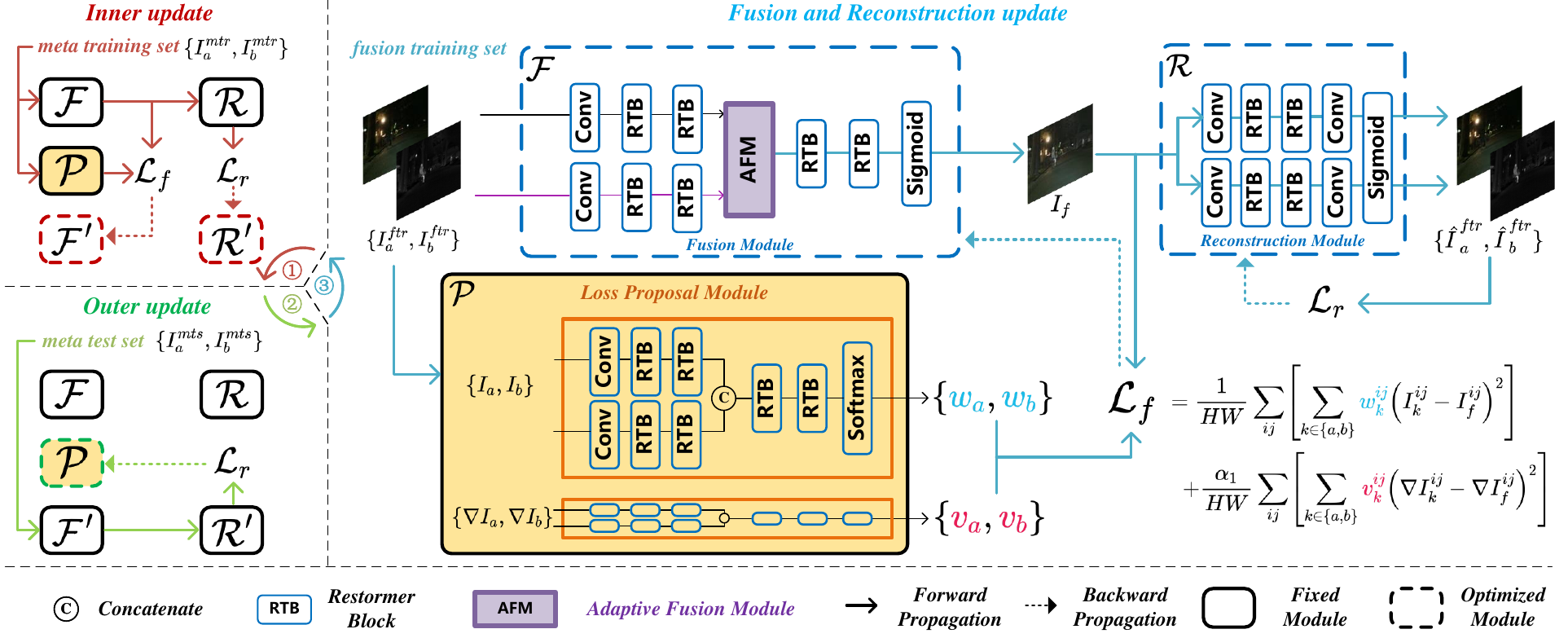}
	\caption{Workflow illustration of ReFusion. The alternating three stages are denoted by red, blue, and green. The inner update, denoted by red, attempts to update the $\mathcal{F}$ using the currently proposed fusion loss. The outer update, denoted by green, updates the $\mathcal{P}$ using the reconstruction loss of the meta-test set. The fusion and reconstruction update stage, denoted by blue, optimizes the $\mathcal{F}$ and $\mathcal{R}$ using $\mathcal{L}_f$ and $\mathcal{L}_r$.}
	\label{fig:Refusion}
\vspace{-1em}
\end{figure*}

\subsection{Meta-learning in Vision}\label{Meta-learning in Vision}
Meta-learning, also known as learning to learn, focuses on developing algorithms to enable models to automatically learn optimal hyperparameters for a given task. It has found widespread applications across diverse domains, including robotics, natural language processing, and computer vision. For example, MAML~\citep{finn2017model} and its variants~\citep{finn2019online,nichol2018first,DBLP:conf/cvpr/QinQPJ23} aim to learn effective model initialization weights, enabling the model to quickly adapt to new tasks with limited examples. Building on MAML~\citep{finn2017model}, Meta-SGD~\citep{li2017meta} further learns the optimal update directions and learning rates for few-shot classification and regression tasks. MW-Net~\citep{shu2019meta} and L2RW~\citep{ren2018learning}, on the other hand, focus on learning the important weights of samples using a small, clean validation set to address learning with noisy data problems. Moreover, studies such as \citep{DBLP:conf/nips/HouthooftCISWHA18,DBLP:conf/nips/AntoniouS19,DBLP:conf/iccv/BaikCKCML21} explore learning the loss function to enhance adaptability and generalizability.

In image fusion,  \citep{DBLP:journals/tip/LiCLCY21} proposed a fusion network capable of outputting images at any desired resolution by utilizing a learning network that predicts filters to scale outputs artificially.
MetaFusion~\citep{zhao2023metafusion} introduced a mutual promotion algorithm that enhances both image fusion and object detection, incorporating an embedding module that integrates semantic features compatible with the fusion features.
\cite{DBLP:conf/mm/LiuLL021} searched for effective network architectures based on fusion mechanisms and the characteristics of source images.
In addition, the continuous learning approach~\citep{9151265}, which focuses on multi-task fusion, is closely associated with meta-learning principles.
In this work, we focus on refining the fusion loss function, employing a loss proposal module trained within a meta-learning framework to predict the parameterized fusion loss.

\subsection{Comparison with Existing Approaches}\label{Comparison with Existing Approaches}
Unlike previous fusion approaches based on meta-learning, our work specifically targets the challenges associated with designing fusion loss functions.
This emphasis represents a concerted effort to refine and automate the design of effective and adaptive fusion loss functions, a process that previously depended heavily on manual intervention.
Our methodology stands out with its learnable parameterized loss function, whose parameters are dynamically proposed by the loss proposal module.
This adaptability allows our loss function to be versatile, suitable for a wide range of scenes and tasks.
Moreover, while existing methods often incorporate a reconstruction module within the framework to aid in training the fusion network, our approach uses reconstruction loss to specifically train the loss proposal module. This is aimed at maximizing the preservation of information from the source images.
This approach does not affect the training of the fusion module itself, thereby enabling a more refined and autonomous learning process for the fusion network.

\section{Method}\label{Method}
In this section, we introduce the framework of our proposed method and its concrete training procedure. 
While our method is capable of accommodating multiple image fusion scenarios, for the sake of clarity, we limit our discussion to the fusion of two source images.
Throughout the remainder of this paper, we denote the two
source images and the fused image as $I_a$, $I_b$ and $I_f$. The reconstructed images corresponding to $I_a$, $I_b$ are denoted as $\hat{I_a}$ and $\hat{I_b}$, respectively.

\subsection{Overview}

As illustrated in Fig.~\ref{fig:Refusion}, ReFusion consists of three components, namely, the fusion module $\mathcal{F}(\cdot)$,
the source reconstruction module $\mathcal{R}(\cdot)$ and the loss proposal module $\mathcal{P}(\cdot)$. Their parameters are denoted respectively as $\theta_\mathcal{F}$, $\theta_\mathcal{R}$, $\theta_\mathcal{P}$. $\mathcal{F'}$ and $\mathcal{R'}$ act as interim updated versions of $\mathcal{F}$ and $\mathcal{R}$, playing a crucial role in the subsequent update of $\mathcal{P}$. The parameters of $\mathcal{F'}$ and $\mathcal{R'}$ are respectively denoted as $\theta_\mathcal{F'}$, $\theta_\mathcal{R'}$.

The learning process of ReFusion consists of three alternating stages, represented by red~\normalsize{\textcircled{\footnotesize{1}}}\normalsize, green~\normalsize{\textcircled{\footnotesize{2}}}\normalsize, and blue~\normalsize{\textcircled{\footnotesize{3}}}\normalsize~in Fig.~\ref{fig:Refusion}. 
Dotted boxes indicate the modules updated in each stage. The learnable parameterized fusion loss is denoted as $\mathcal{L}_f$ and its explicit formulation will be given in the next subsection. 
Unlike previous work with source image reconstruction, the reconstruction loss, denoted by $\mathcal{L}_r$, does not play a role in updating the fusion module.

\subsection{Learnable Fusion Loss}\label{loss}
The learnable fusion loss function consists of two independent pairs of outputs generated by the loss proposal module $\mathcal{P}$.  
The pairs $\{ w_{a}, w_{b} \} $ and $\{v_{a}, v_{b} \}$ are computed by $\mathcal{P}$ using both the source images and their gradients, \ie, $\mathcal{P}(I_{a}, I_{b},\nabla I_{a},\nabla I_{b})$.
Here, $w$ and $v$ are parameters that define the weights for the intensity and gradient preferences of the source images, respectively, ensuring that $w_{a}^{ij}+w_{b}^{ij}=1$ and $v_{a}^{ij}+v_{b}^{ij}=1$ for all pixels.
In this context, $\nabla$ denotes the Sobel operator, commonly utilized to extract gradients in image fusion applications.~\citep{zhao2023cddfuse,DBLP:conf/cvpr/LiuFHWLZL22,tang2022image}.
The specific formulation of the learnable fusion loss is defined as:
\begin{align}
    \mathcal{L}_f&=\mathcal{L}_f^{int}+\alpha_1 \mathcal{L}_f^{grad},\\
    \mathcal{L}_f^{int}&=\frac{1}{H W} \sum_{i j}\left[\sum_{k \in\{a, b\}} w_k^{ij}\left(I_k^{i j}-I_f^{i j}\right)^2\right],\label{eq2}\\
    \mathcal{L}_f^{grad}&=\frac{1}{H W} \sum_{i j}\left[\sum_{k \in\{a, b\}} v_k^{ij}\left(\nabla I_k^{i j}-\nabla I_f^{i j}\right)^2\right],
\end{align}
where $\alpha_1$ serves as a scaling parameter, $\mathcal{L}_f^{int}$ and $\mathcal{L}_f^{grad}$ denote the intensity loss and gradient loss.  
The weights $w_{a}$ and $w_{b}$ determine the loss function's preference for the intensity information from each source image. $v_{a}$ and $v_{b}$ reflect the emphasis on the gradient information.
These parameters enable the fusion process to selectively extract and incorporate pertinent information from the source images. Variations in $\theta_\mathcal{P}$ lead to different configurations of $w$ and $v$, thus affecting the characteristics of the fusion loss.
$\theta_\mathcal{P}$ is updated based on the reconstruction loss during the in-step update process, as detailed in the subsequent section.

The reconstruction loss, comprising both the intensity and gradient losses, is specifically formulated as follows:
\begin{align}
    \mathcal{L}_r&=\mathcal{L}_r^{int}+\alpha_2 \mathcal{L}_r^{grad},\\
    \mathcal{L}_r^{int}&=\frac{1}{H W} \sum_{i j} \max _{k \in\{a, b\}}\left[\left(I_k^{i j}-\hat{I}_k^{i j}\right)^2\right],\label{eq5}\\
    \mathcal{L}_r^{grad}&=\frac{1}{H W} \sum_{i j} \max _{k \in\{a, b\}}\left[\left(\nabla I_k^{i j}-\nabla \hat{I}_k^{i j}\right)^2\right],\label{eq6}
\end{align}	
where $\alpha_2$ is a scaling parameter. 
The $Max(\cdot)$ function, operating at the image pixel level, aims to balance the degree of the reconstruction among different source images and promote stability during the training phase. 
If the reconstruction loss of the source images converges to significantly different values, the fused image may closely resemble the more accurately reconstructed source image, leading to fusion failure.
This will be further discussed in the ablation studies section.

\subsection{Fusion from Reconstruction}\label{FFR}
{To finetune the guiding abilities of the loss proposal module $\mathcal{P}$ for fusion tasks, non-overlapping subsets with size $\emph{M}$ are created for each epoch. The meta-training set $\{I_a^{mtr}, I_b^{mtr}\}$ and the meta-testing set $\{I_a^{mts}, I_b^{mts}\}$ are randomly sampled from the fusion training set $\{I_a^{ftr}, I_b^{ftr}\}$.
Samples drawn from these subsets are sequentially introduced into the model at specific stages of the training process.
The training process is divided into three stages. In the first two stages, the inner and outer updates apply and optimize $\mathcal{P}$, respectively. In the third stage, the updates focus on the fusion and reconstruction processes, using the losses $\mathcal{L}_f$ and $\mathcal{L}_r$ to enhance the performance of $\mathcal{F}$ and $\mathcal{R}$.}

\subsubsection{Inner Update}
{During the inner update, we attempt to update $\mathcal{F}$ using the fusion loss defined by the current state of the loss proposal module $\mathcal{P}$. 
The inner update is primarily aimed at obtaining the intermediate parameters $\theta_\mathcal{F'}$ and $\theta_\mathcal{R'}$ for the subsequent update of $\theta_\mathcal{P}$.}
The flow of this stage is represented by the color red~\normalsize{\textcircled{\footnotesize{1}}}\normalsize~in the framework diagram in Fig.~\ref{fig:Refusion}.
At this stage, images from the meta-training set $\{I_a^{mtr}, I_b^{mtr}\}$ are fed into the model, and $\mathcal{F}$ undergoes a single update via gradient descent:
\begin{equation}
    \label{equ1}
    \small
    \theta_{\mathcal{F'}}=\theta_\mathcal{F}-\eta_\mathcal{F'} \frac{\partial \mathcal{L}_f\left(I_a^{mtr}, I_b^{mtr}, I_f^{mtr} ; \theta_\mathcal{P}\right)}{\partial \theta_\mathcal{F}},
\end{equation}
where $\theta_\mathcal{P}$ represents the parameters of $\mathcal{P}$, and the outputs of $\mathcal{P}$ are the parameters of the learnable fusion loss.
In the update equation, $\eta_\mathcal{F'}$ is the step size used for updating the fusion module. The module $\mathcal{F'}$ acts as a temporary substitute for $\mathcal{F}$ and updates the parameters of $\mathcal{F}$ following a single update step. 
$\theta_\mathcal{F}$, the parameters of $\mathcal{F}$, are not updated in this process.
Analogously, $\mathcal{R'}$ undergoes a one-step update relying on the parameters $\theta_\mathcal{R}$ of the current source reconstruction module $\mathcal{R}$:
\begin{equation}\label{equ2}
    \small
    \theta_{\mathcal{R'}}=\theta_\mathcal{R}-\eta_\mathcal{R'} \frac{\partial \mathcal{L}_r\left(I_a^{mtr}, I_b^{mtr}, \hat{I}_a^{mtr}, \hat{I}_b^{mtr}\right)}{\partial \theta_\mathcal{R}}.
\end{equation}
{The inner update modifies the parameters $\theta_{\mathcal{F'}}$ and $\theta_{\mathcal{R'}}$. It also preserves the computation graph of $\theta_{\mathcal{F'}}$ in relation to $\theta_{\mathcal{P}}$. This preservation facilitates the optimization of $\theta_{\mathcal{P}}$ during the outer update.}

\subsubsection{Outer Update}
{The primary objective of the outer update is to evaluate the current fusion guidance capability of $\mathcal{P}$ and refine it, that is, to enhance the effectiveness of the loss function $\mathcal{L}_f$ in directing the fusion module $\mathcal{F}$. }
This stage is represented by the color green~\normalsize{\textcircled{\footnotesize{2}}}\normalsize~in the framework illustration. 
The modules $\mathcal{F'}$ and $\mathcal{R'}$, derived from the inner update, represent the current instructional capabilities of $\mathcal{P}$.
Ideally, the optimal fusion loss would yield a fused image from which the source images can be reconstructed more easily.
Therefore, during this step, the meta-test set $\{I_a^{mts}, I_b^{mts}\}$ is utilized. 
The parameters $\theta_{\mathcal{P}}$ are updated based on the reconstruction loss $\mathcal{L}_r$, computed by $\mathcal{F'}$ and $\mathcal{R'}$:
\begin{equation}\label{equ3}
\small
	\theta_\mathcal{P}=\theta_\mathcal{P}-\eta_\mathcal{P} \frac{\partial \mathcal{L}_r\left(I_a^{mts}, I_b^{mts}, \hat{I}_a^{mts}, \hat{I}_b^{mts}\right)}{\partial \theta_\mathcal{P}},
\end{equation}
where the gradient $\partial \mathcal{L}_r/\partial \theta_\mathcal{P}$ can be calculated as:
\begin{equation}
\small
	\frac{\partial \mathcal{L}_r}{\partial \theta_\mathcal{P}}=\frac{\partial \mathcal{L}_r}{\partial \theta_\mathcal{F'}} \!*\!\left(-\eta_\mathcal{F'} \frac{\partial^2 \mathcal{L}_f\left(I_a^{mtr}, I_b^{mtr}, I_f^{mtr} ; \theta_\mathcal{P}\right)}{\partial \theta_\mathcal{F} \partial \theta_\mathcal{P}}\right).
\end{equation}
{Eq.~(\ref{equ3}) holds because $\mathcal{L}_r$ is computed from $\hat{I}_a^{mts}$ and $\hat{I}_b^{mts}$, which are reconstructed by $I_f^{mts}$, and $I_f^{mts}$ is related to $\theta_\mathcal{F'}$. By preserving the computation graph of $\theta_\mathcal{F'}$ in relation to $\theta_\mathcal{P}$ during the inner update, $\theta_\mathcal{P}$ can be successfully optimized by $\mathcal{L}_r$.}
The updated $\mathcal{P}$ module excels at formulating enhanced fusion loss functions, enabling the fusion module to more effectively integrate comprehensive information from the source images into the fused image.

\subsubsection{Fusion and Reconstruction Update}
The cyclical process, which involves several iterations of inner and outer updates, 
establishes a dynamic and efficient mechanism for the enhancement of $\mathcal{P}$ with respect to the current state of $\mathcal{F}$. 
Subsequently, the refined $\mathcal{P}$ is employed to further advance the training of $\mathcal{F}$.
At this juncture, indicated by the color blue~\normalsize{\textcircled{\footnotesize{3}}}\normalsize~in the diagrams, the images of the fusion training set $\{I_a^{ftr}, I_b^{ftr}\}$ are processed. $\mathcal{F}$ and $\mathcal{R}$ undergo updating through the respective application of the fusion loss $\mathcal{L}_f$ and the reconstruction loss $\mathcal{L}_r$:
\setlength{\abovedisplayskip}{0.0cm}
\begin{align}
\theta_{\mathcal{F}}=\theta_\mathcal{F}-\eta_\mathcal{F} \frac{\partial \mathcal{L}_f\left(I_a^{ftr}, I_b^{ftr}, I_f^{ftr} ;\theta_{\mathcal{P}}\right)}{\partial \theta_\mathcal{F}},\label{equ4}\\
\theta_{\mathcal{R}}=\theta_\mathcal{R}-\eta_\mathcal{R} \frac{\partial \mathcal{L}_r\left(I_a^{ftr}, I_b^{ftr}, \hat{I}_a^{ftr}, \hat{I}_b^{ftr}\right)}{\partial \theta_\mathcal{R}}.\label{equ5}
\end{align}
The training of the fusion framework proceeds in a sequence of alternating stages, each tailored to optimally calibrate the fusion loss for the network's current state.
This systematic alternation ensures the fusion network consistently employs the most suitable fusion loss at each stage of its development.
The culmination of this process is the development of a highly effective fusion network, meticulously optimized for superior performance. 
The entire training protocol is detailed in the algorithm outlined in Algorithm~\ref{alg:moal}.

\begin{algorithm}[t]
	\caption{ReFusion Training Algorithm}
	\label{alg:moal}
	\begin{algorithmic}[1]
		\Require 
		Training set $\{I_a^{ftr}, I_b^{ftr}\}$ with size $\emph{N}$.
		\Ensure  
		{Thoroughly trained $\theta_{\mathcal{F}}$, $\theta_{\mathcal{R}}$, $\theta_{\mathcal{P}}$.}
		\State Initialize $\theta_{\mathcal{F}}$, $\theta_{\mathcal{R}}$, $\theta_{\mathcal{P}}$.
		\For {$epoch=1$ \text{\textbf{to}} $\textbf{\emph{L}}$}
		\State Sample meta-training set $\{I_a^{mtr}, I_b^{mtr}\}$ and 
        \Statex \qquad meta-test set $\{I_a^{mts}, I_b^{mts}\}$ from $\{I_a^{ftr},I_b^{ftr}\}$.

		\For {$step=1$ \text{\textbf{to}} $\textbf{\emph{M}}$}
		\Statex \qquad \quad \% \textit{Inner update: attempt to apply $\mathcal{P}$}.
		\State Sample $(I_a^{mtr},I_b^{mtr})$, get  $(I_f^{mtr},\hat{I}_a^{mtr},\hat{I}_b^{mtr})$. 
		\State Compute $\theta_{\mathcal{F'}}$ and $\theta_{\mathcal{R'}}$ by Eq.~(\ref{equ1}) and Eq.~(\ref{equ2}).
		\Statex \qquad \quad \% \textit{Outer update: optimize $\mathcal{P}$}.
		\State Sample $(I_a^{mts},I_b^{mts})$, get $(I_f^{mts},\hat{I}_a^{mts}, \hat{I}_b^{mts})$. 
		\State Update $\theta_{\mathcal{P}}$ by Eq.~(\ref{equ3}).
		\EndFor
		\For {$step=1$ \text{\textbf{to}} $\textbf{\emph{N}}$}
		\Statex \qquad \quad \% \textit{Fusion and Reconstruction update: optimize} 
        \Statex \qquad \quad \textit{$\mathcal{F}$ and $\mathcal{R}$}.
		\State Sample $(I_a^{ftr},I_b^{ftr})$, get $(I_f^{ftr},\hat{I}_a^{ftr},\hat{I}_b^{ftr})$.
		\State Update $\theta_{\mathcal{F}}$ and $\theta_{\mathcal{R}}$ by Eq.~(\ref{equ4}) and Eq.~(\ref{equ5}).
		\EndFor
		\EndFor                        
	\end{algorithmic}

\end{algorithm}

\subsection{Network Architecture}

\begin{figure}[t]
	\centering
	\includegraphics[width=\linewidth]{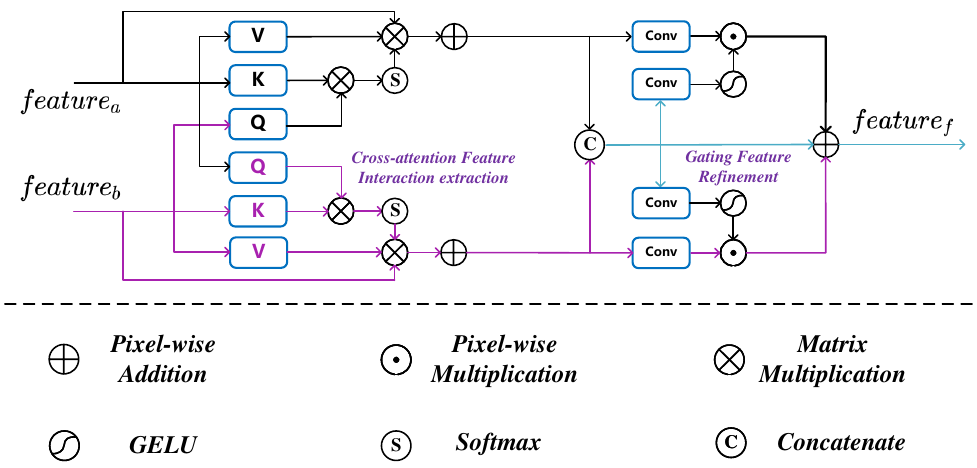}
	\caption{
 The structure of the Adaptive Fusion Module (AFM), which consists cross-attention interactive feature extraction and gating features refine.}
 \label{fig:AFM}

\end{figure}

The structure of ReFusion is shown in Fig.~\ref{fig:Refusion}. The basic unit of ReFusion is constituted by Restormer Block (RTB)~\citep{zamir2022restormer}. RTB utilizes channel-dimensional self-attention operations to extract features from images. Further details on its structure can be found in the original paper~\citep{zamir2022restormer}.
Within the fusion module, the RTB separates the unique information from both the source and composite fusion images. This information is derived from the fusion features, which are produced by the Adaptive Fusion Module (AFM). The AFM initially employs cross-attention to enable interactive feature extraction from the source images. After concatenating these features, it uses a gating mechanism to refine and integrate the unique features of the source images into the fusion features.
The detailed structure of AFM is shown in Fig.~\ref{fig:AFM}.
{The $Softmax(\cdot)$ layer in the LPM structure ensures that the weights $w_{a}^{ij}+w_{b}^{ij}=1$ and $v_{a}^{ij}+v_{b}^{ij}=1$ for each pixel. This design ensures that the fused image will approximate the source images even under suboptimal initialization conditions, making it easier to be corrected by reconstruction loss in subsequent training. As a result, there is no need for additional initialization or pre-training of the network before training begins.}

\subsection{Theoretical Analysis}
To gain a deeper understanding of the weighting scheme of the proposed module $\mathcal{P}$, we begin by analyzing the optimization process of the $\mathcal{P}$ component which is responsible for generating the weights $\{w_a, w_b\}$, denoted as $\theta_{\mathcal{P}_1}$. For the sake of simplicity, we rephrase Eq.~(\ref{eq2}) as follows,

\begin{equation}
    \begin{aligned}
		\mathcal{L}_f^{int} = &[{w}_a\odot({I}_a-{I}_f)\odot({I}_a-{I}_f)   \\
			&+ {w}_b\odot({I}_b-{I}_f)\odot({I}_b-{I}_f) ] \times \frac{1}{HW} \\
			= & [\mathcal{P}({I}_a,{I}_b;\theta_{\mathcal{P}_1})  \odot({I}_a-\mathcal{F}_{\theta_{\mathcal{F}}}({I}_a,{I}_b)) 
            \\ &\quad\odot({I}_a-\mathcal{F}_{\theta_{\mathcal{F}}}({I}_a,{I}_b))  \\
			&  +(1-\mathcal{P}({I}_a,{I}_b;\theta_{\mathcal{P}_1}))\odot({I}_b-\mathcal{F}_{\theta_{\mathcal{F}}}({I}_a,{I}_b)) \\ & \quad\odot({I}_b-\mathcal{F}_{\theta_{\mathcal{F}}}({I}_a,{I}_b)) ] \times \frac{1}{HW}, 
   \end{aligned}
\end{equation}
where $w \in \mathbb{R}^{H\times W}$, $I_a, I_b \in \mathbb{R}^{H\times W}$, and $\odot$ represents the pixel-wise multiplication operation. Let $\Omega^{'}=\{\theta_{\mathcal{F}^{'}},\theta_{\mathcal{R}^{'}}\}$, then we can get:
\begin{equation}
    \begin{aligned}
		  \theta_{\mathcal{P}_1} =& \theta_{\mathcal{P}_1} -\eta_{\mathcal{P}} \frac{\partial \mathcal{L}_r^{mts}(\Omega^{'}(\theta_{\mathcal{P}_1}))}{\partial \theta_{\mathcal{P}_1}} \\
            = &\theta_{\mathcal{P}_1} - \eta_{\mathcal{P}} \frac{\partial \mathcal{L}_r^{mts}(\Omega^{'}(\theta_{\mathcal{P}_1}))} {\partial \Omega^{'}}  \frac{\partial \Omega^{'}(\theta_{\mathcal{P}_1})}{\partial \theta_{\mathcal{P}_1}} \\
			 = & \theta_{\mathcal{P}_1} - \eta_{\mathcal{P}}\eta_{\mathcal{F}^{'}} \underbrace{\frac{\partial \mathcal{L}_r^{mts}(\Omega^{'}(\theta_{\mathcal{P}_1}))} {\partial \Omega^{'}}}_{(\mathrm{a})} \times  \frac{\partial \mathcal{P}({I}_a,{I}_b;\theta_{\mathcal{P}_1})}{\theta_{\mathcal{P}_1}} \\& \times\underbrace{[({I}_a - \frac{\partial \mathcal{F}_{\theta_{\mathcal{F}}}}{\partial{\theta_{\mathcal{F}}}})\odot({I}_a - \frac{\partial \mathcal{F}_{\theta_{\mathcal{F}}}}{\partial{\theta_{\mathcal{F}}}})}_{(\mathrm{b})} \\ & \quad \underbrace{-({I}_b - \frac{\partial \mathcal{F}_{\theta_{\mathcal{F}}}}{\partial{\theta_{\mathcal{F}}}})\odot({I}_b - \frac{\partial \mathcal{F}_{\theta_{\mathcal{F}}}}{\partial{\theta_{\mathcal{F}}}}) ]}_{(\mathrm{b})} \\
		   = & \theta_{\mathcal{P}_1}-\eta_{\mathcal{P}}\eta_{\mathcal{F}^{'}} \mathbf{G}\times \frac{\partial \mathcal{P}({I}_a,{I}_b;\theta_{\mathcal{P}_1})}{\theta_{\mathcal{P}_1}},
    \end{aligned}
\end{equation}
where $\mathbf{G}$ is the inner product of term (a) and term (b), which are the gradient calculated on reconstruction loss using a \textbf{meta-testing set} and the gradient computed on the fusion loss using a \textbf{meta-training set}, respectively. It can reflect the similarity between the gradient of the fusion loss and the reconstruction loss. Similarly, for the component $\theta_{\mathcal{P}_2}$ responsible for generating weights $\{v_a, v_b\}$, we can draw a similar conclusion. Therefore, it is evident that the optimization of the module $\theta_{\mathcal{P}}$ is guided by the reconstruction loss, with the objective of preserving as much information as possible from each image during the fusion process.

\section{Experiment}
\subsection{Setup}
The fusion experiments include infrared-visible image fusion (IVIF), medical image fusion (MIF), multi-focus image fusion (MFIF), and multi-exposure image fusion (MEIF). 
For these tasks, details on the datasets used and the comparison methods will be provided in the subsequent sections. 
The evaluation metrics adopted can be referenced in~\citep{zhang2021benchmarking,DBLP:journals/inffus/LiuWCLC20,DBLP:journals/inffus/MaML19}.
Throughout the training phase for all tasks, we standardize the number of epochs $\textbf{\emph{L}}$ to 50, with $\textbf{\emph{M}}$ and $\textbf{\emph{N}}$ specified as 600 and 1600, respectively.
In the training process, the input images are cropped to $128\times128$ patches.
We utilize the Adam optimizer, initiated with a learning rate of 1e-4.
The batch size is configured to 4, and the hyperparameters $\alpha_1$ and $\alpha_2$ are uniformly set to 1.
Computational experiments are executed on a PC with a single NVIDIA GeForce RTX 3090 GPU.

The RGB input is initially transformed into the YCrCb color space to facilitate the fusion of the luminance (Y) channel. The chrominance channels, Cr and Cb, are then fused as follows:
\begin{equation}
\small
\mathbf{C}_f=\frac{\mathbf{C}_1\left(\left|\mathbf{C}_1-\tau\right|\right)+\mathbf{C}_2\left(\left|\mathbf{C}_2-\tau\right|\right)}{\left|\mathbf{C}_1-\tau\right|+\left|\mathbf{C}_2-\tau\right|},
\end{equation}
where $\mathbf{C}_1$ and $\mathbf{C}_2$ represent the Cr or Cb channels data from the source images,  and $\mathbf{C}_f$ denotes the fused chrominance data. The constant $\tau$ is set to 128. Following the completion of fusion within the YCrCb space, the data is reconverted back to RGB space. This process effectively transforms the multi-channel fusion task into a single-channel operation.

\begin{figure*}[t]
        \vspace{1em}
	\centering
	\includegraphics[width=\linewidth]{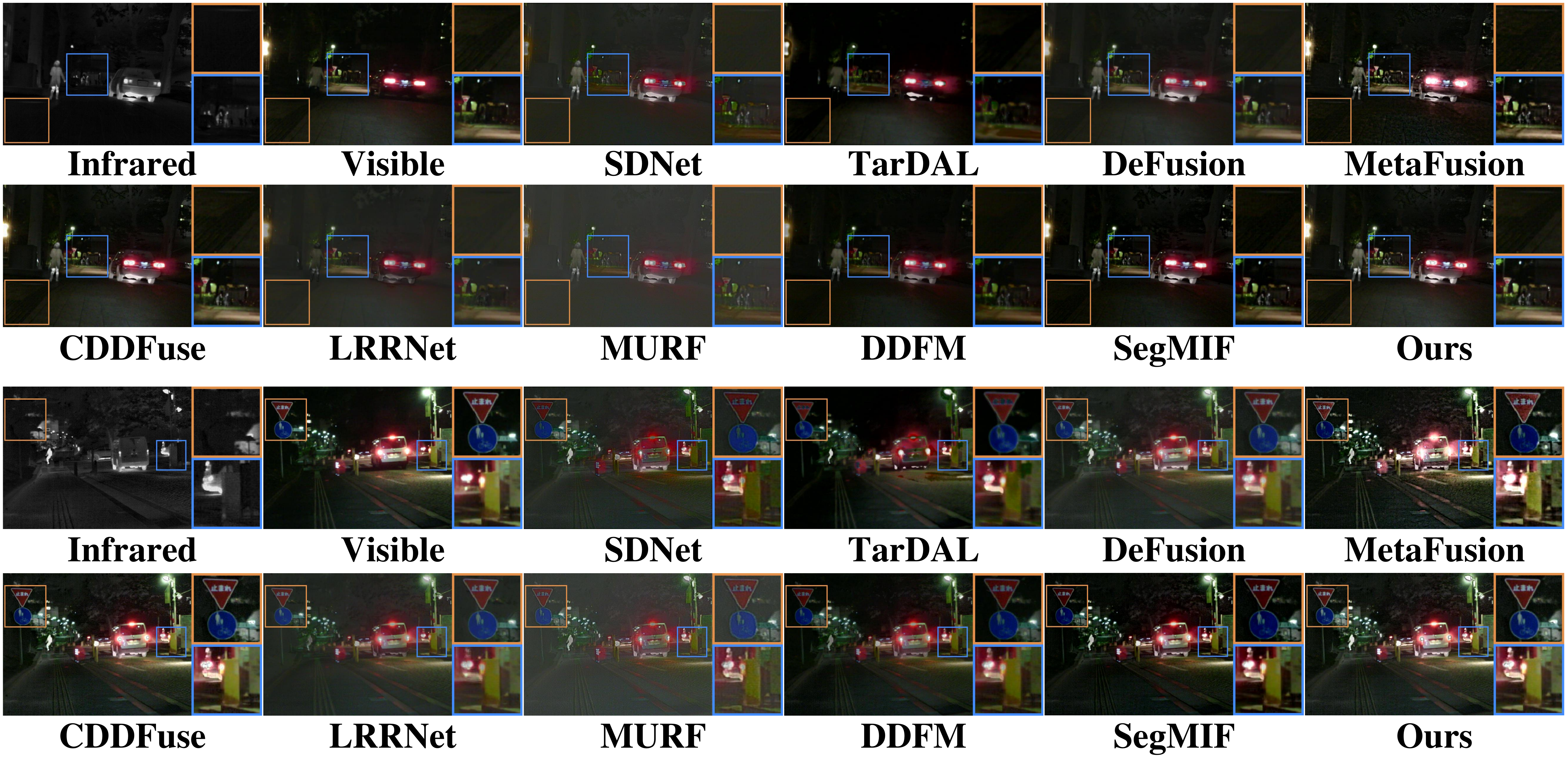}
	\caption{Visual comparison for ``00710N'' and ``00906N'' in MSRS dataset for IVIF.}
	\label{fig:IVIF}
    
\end{figure*}
\begin{table*}[t]
	\centering
        \caption{Quantitative results of infrared-visible image fusion. The best and second-best values are \textbf{highlighted} and \underline{underlined}.}
	\resizebox{\linewidth}{!}{
		\begin{tabular}{lccccccccccccc}
			\toprule
			&\multicolumn{6}{c}{\textbf{Infrared-visible Image Fusion on MSRS}}                              &&                              \multicolumn{6}{c}{\textbf{Infrared-visible Image Fusion on RoadScene}}                               \\
		Methods&  EN $\uparrow$   &   SD $\uparrow$   &  SF $\uparrow$   &  AG $\uparrow$  & SCD $\uparrow$  & SSIM  $\uparrow$  & &                                    EN $\uparrow$   &   SD $\uparrow$   &  SF $\uparrow$   &  AG $\uparrow$  & SCD $\uparrow$  & SSIM  $\uparrow$  \\ \midrule
			SDNet \citep{DBLP:journals/ijcv/ZhangM21}  & 5.25  & 17.35  & 8.67  & 2.67  & 0.99  & 0.36 & &  7.10  &  38.93 & 12.66  & 4.91  &  1.42 & \underline{0.72}  \\
			TarDAL~\citep{DBLP:conf/cvpr/LiuFHWLZL22}  & 5.28  & 25.22  & 5.98  & 1.83  & 0.71  & 0.24 & &  7.11  & 43.13  & 9.60  & 3.43  & 1.47  &  0.71 \\
			DeFusion~\citep{Liang2022ECCV}  & 6.46  & 37.63  & 8.60  & 2.80  & 1.35  & 0.47 & &  7.23  & 43.38  & 9.38  & 3.68  & 1.68  & 0.69  \\
			MetaFusion~\citep{zhao2023metafusion}  & 5.65  & 24.97  &  9.99 & 3.40  & 1.14  & 0.18  &&  6.70  & 28.59  & 12.43 & 4.72 &0.93  & 0.66 \\
			CDDFuse~\citep{zhao2023cddfuse} & \underline{6.70}  & \textbf{43.38}  & \underline{11.56}  & \underline{3.73}  & \underline{1.62}  & \underline{0.50}  & & \underline{7.28}  & \underline{44.43}  & \underline{14.10} &  \underline{5.13} & \underline{1.69}  & 0.70  \\
			LRRNet~\citep{li2023lrrnet} & 6.19  & 31.78  & 8.46  &  2.63 &  0.79 &  0.22 &&  7.02  & 37.15  & 9.87  & 3.69  & 1.56  & 0.62  \\
			MURF~\citep{xu2023murf} & 5.04 &  16.37 & 8.31 & 2.67  & 0.86  & 0.29  &&   6.80 & 31.10  & 12.96 & 4.88  & 1.16  & 0.71  \\
			DDFM~\citep{zhao2023ddfm} & 6.19 &  29.26 & 7.44  & 2.51  & 1.45  & 0.45 & &   7.09 & 38.46  & 9.11  & 3.46  & 1.67  & 0.71  \\
			SegMIF~\citep{Liu_2023_ICCV} & 5.95 &  37.28 & 11.10 & 3.47  & 1.57  & 0.35 & &   7.15 & 42.11  & 12.31  & 4.54  & 1.57  & 0.59  \\
			ReFusion (Ours) &  \textbf{6.71} & \underline{42.89} & \textbf{11.64} & \textbf{3.80} & \textbf{1.67} & \textbf{0.51} &&  \textbf{7.28} & \textbf{45.65} &  \textbf{14.20} & \textbf{5.23} & \textbf{1.70} &  \textbf{0.72} \\
			\bottomrule
	\end{tabular}}
	
	\label{tab:IVIF}%
\end{table*}%

\subsection{Infrared-Visible Image Fusion}
Our study employs the MSRS~\citep{DBLP:journals/inffus/TangYZJM22} and RoadScene~\citep{xu2020aaai} datasets for infrared-visible image fusion, where MSRS comprises 1083 training pairs and 361 test pairs. We train and test our model on MSRS, then assess its generalization on 50 RoadScene~\citep{xu2020aaai} pairs. 
We compare our model  with several state-of-the-art methods, including SDNet~\citep{DBLP:journals/ijcv/ZhangM21}, TarDAL~\citep{DBLP:conf/cvpr/LiuFHWLZL22}, DeFusion~\citep{Liang2022ECCV}, MetaFusion~\citep{zhao2023metafusion}, CDDFuse~\citep{zhao2023cddfuse}, LRRNet~\citep{li2023lrrnet}, MURF~\citep{xu2023murf}, DDFM~\citep{zhao2023ddfm}, and SegMIF~\citep{Liu_2023_ICCV}. 
{Focusing on integrating diverse modality images, performance is evaluated using metrics such as entropy (EN), standard deviation (SD), spatial frequency (SF), average gradient (AG), sum of correlation differences (SCD), and structural similarity index metric (SSIM).}
Qualitative analysis, as shown in Fig.~\ref{fig:IVIF}, reveals that ReFusion effectively preserves the details of visible images, including gradient information and edge textures, especially in highlighted areas like ground textures and road signage. It also enhances the thermal radiation features of infrared images, capturing intricate details of distant figures, such as the structures of vehicles and pedestrians. This capability is particularly beneficial for nighttime driving, where enhanced visual information aids in safer navigation and improved situational awareness. 
The ability to clearly see and distinguish thermal signatures alongside traditional visual cues helps in identifying potential hazards and obstacles in low-light conditions, making it invaluable for advanced driver-assistance systems (ADAS) and autonomous vehicle technologies.
Quantitative assessment in Table~\ref{tab:IVIF} shows that our method leads in SD, SF, and AG, indicating superior texture preservation. Additionally, top scores in EN, SCD, and SSIM reflect its efficacy in retaining crucial, complementary data from source images for a richer scene depiction. 
Both qualitative and quantitative results underscore the superior capability of ReFusion in fusing images from different modalities in IVIF.

\begin{figure*}[t]
	\centering

	\includegraphics[width=\linewidth]{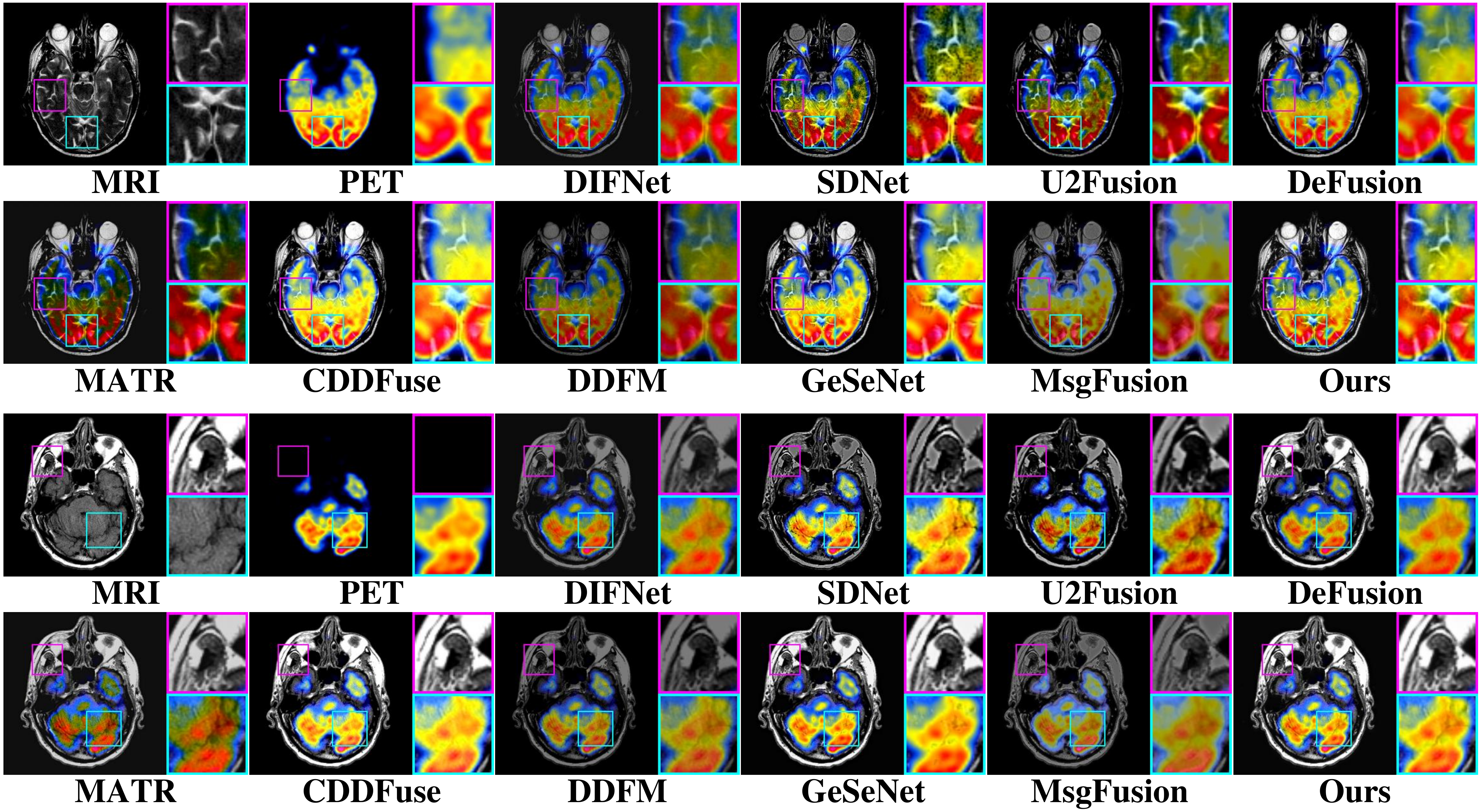}
	\caption{Visualization of the MRI-PET case in Harvard dataset for MIF.}
	\label{fig:MIF}
\end{figure*}

\subsection{Medical Image Fusion}

In our medical image fusion research, we use 158 and 50 image pairs from the Harvard dataset\footnotemark[1] for training and testing, respectively.
\footnotetext[1]{http://www.med.harvard.edu/AANLIB/home.html}
Our approach is compared against nine state-of-the-art methods in the field, including DIF-Net~\citep{jung2020unsupervised}, SDNet~\citep{DBLP:journals/ijcv/ZhangM21}, U2Fusion~\citep{9151265}, DeFusion~\citep{Liang2022ECCV}, MATR~\citep{tang2022matr}, CDDFuse~\citep{zhao2023cddfuse}, DDFM~\citep{zhao2023ddfm}, GeSeNet~\citep{li2023gesenet} and MsgFusion~\citep{wen2023msgfusion}. The evaluation metrics are the same as those used in IVIF. 
ReFusion excels at preserving key histological and functional information from MRI and PET scans, as highlighted in Fig.~\ref{fig:MIF}. This dual preservation is crucial for accurate diagnosis and disease tracking. The ability to retain and enhance such critical data supports medical professionals in making more informed decisions and improves the accuracy of diagnosing and monitoring various conditions. This is particularly significant in oncology, where precise imaging can significantly impact treatment planning and patient outcomes.
The quantitative result presented in Table~\ref{tab:MIF} shows that our method achieves the highest scores across most metrics, despite a slight underperformance in SSIM compared to DeFusion~\citep{Liang2022ECCV}. This affirms its efficacy in integrating essential details from the source images.

\begin{table}[h]
\centering
\caption{Quantitative results of MIF. \textbf{Boldface} and \underline{underline} show the best and second-best values, respectively.} 
\resizebox{\linewidth}{!}{
    \begin{tabular}{lcccccccc}
        \toprule
        \multicolumn{7}{c}{\textbf{Medical Image Fusion on Harvard Dataset}}                                 \\
    Methods&        EN $\uparrow$       &        SD $\uparrow$         &        SF $\uparrow$                &       AG $\uparrow$       &       SCD $\uparrow$       &       SSIM $\uparrow$        \\ \midrule
        DIF-Net~\citep{jung2020unsupervised} &       \underline{4.85}       &       52.33       &       16.53       &       4.68       &       1.14       &       0.26       \\
        SDNet~\citep{DBLP:journals/ijcv/ZhangM21}   &       4.23       &       56.79       &       \underline{28.33}        &       6.98       &       1.10       &       0.58        \\
        U2Fusion~\citep{9151265}  &       4.18       &       52.95       &       22.02        &       5.90       &       1.05       &       0.54 \\
        DeFusion~\citep{Liang2022ECCV}   &       4.30       &       63.95       &  21.15   &       5.44       &       1.19       &       \textbf{0.76}  \\
        MATR~\citep{tang2022matr} &       4.62       &       53.35       &       20.52        &       5.75       &       0.40       &       0.26 \\
        CDDFuse~\citep{zhao2023cddfuse}  &       4.41       &       \underline{72.18}       &       28.08        & 7.26 &       \underline{1.59}       & 0.75 \\
        DDFM~\citep{zhao2023ddfm}   & 4.14 & 59.12 &       19.12        &       4.86       & 1.36 &       0.75     \\
        GeSeNet~\citep{li2023gesenet}   & 4.76    & 70.83    &  28.15   &  \underline{7.54}    &  1.54    & 0.51\\
        MsgFusion~\citep{wen2023msgfusion}   & 4.36   & 69.42    & 27.98    &   6.68   &  1.18    & 0.51\\
        ReFusion (Ours)   &  \textbf{5.09}   &  \textbf{72.38}   & \textbf{28.35} &  \textbf{7.60}   &  \textbf{1.62}   &  \underline{0.75}   \\  \bottomrule
    \end{tabular}}
\label{tab:MIF}
\end{table}

\begin{figure*}[t]
	\centering
	\includegraphics[width=\linewidth]{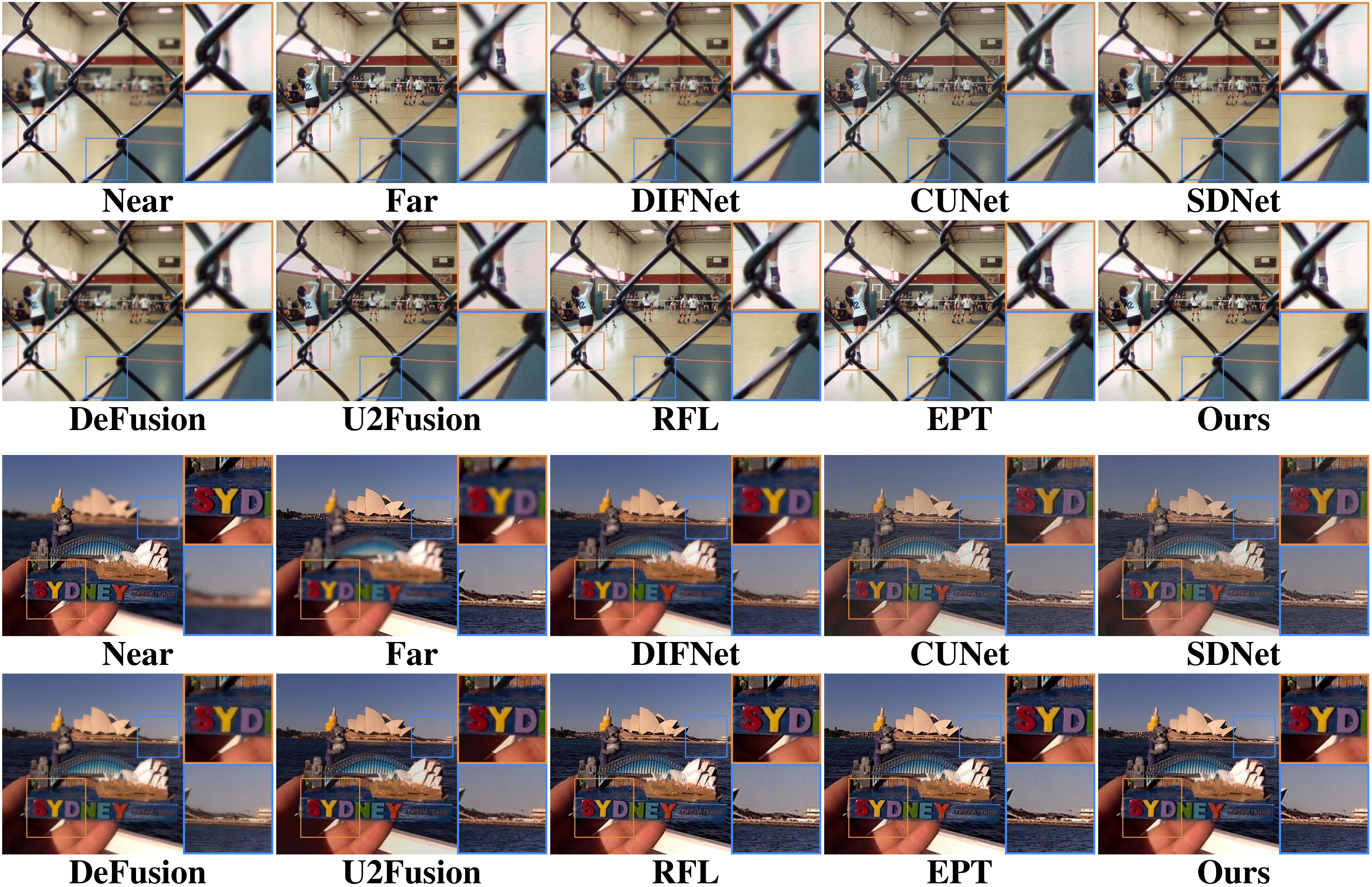}
	\caption{Visualization of the ``Lytro\_05'' and ``Lytro\_14'' in Lytro dataset for MFIF.}
	\label{fig:MFIF}
\vspace{-1em}
\end{figure*}
\begin{table*}[t]
	\centering
	\caption{Quantitative comparison of MFIF. The best and second-best values are \textbf{highlighted} and \underline{underlined}.}
	\label{tab:MFIF}%
	\resizebox{\linewidth}{!}{
		\begin{tabular}{lccccccccccccc}
			\toprule
			& \multicolumn{6}{c}{\textbf{Multi-focus Image Fusion on RealMFF Dataset}}                             & &                              \multicolumn{6}{c}{\textbf{Multi-focus Image Fusion on Lytro Dataset}}                               \\
		Methods&  EN $\uparrow$   &   SD $\uparrow$   &  SF $\uparrow$   &  VIF $\uparrow$  & $Q_{CB}$ $\uparrow$  & $Q_{NCIE}$  $\uparrow$  &                                     &  EN $\uparrow$   &   SD $\uparrow$   &  SF $\uparrow$   &  VIF $\uparrow$  & $Q_{CB}$ $\uparrow$  & $Q_{NCIE}$  $\uparrow$    \\ \midrule
			DIFNet~\citep{jung2020unsupervised}  & 6.85  & 48.87  & 10.98  & 0.86  & 0.65  & 0.83  &    & 7.43  &  52.52 & 11.47  & 0.73  &  0.59 & 0.83  \\
			CUNet~\citep{deng2020deep}  & 6.58  & 37.13  & 13.52  & 0.74  & 0.53  & 0.82  &   & 7.25  & 45.78  & 15.54  & 0.71  & 0.57  &  0.82 \\
			SDNet~\citep{DBLP:journals/ijcv/ZhangM21}  & 6.81  & 49.93  & \underline{15.58}  & 0.92  & 0.72  & 0.83  &   & 7.47  & 55.25  & 16.88  & 0.84  & 0.65  & 0.83  \\
			U2Fusion~\citep{9151265}  & 6.60  & 47.07  &  14.35 & 0.92  & 0.60  & 0.82  &   & 7.30  & 51.95  & 14.83 & 0.83  & 0.64  &0.82  \\
			DeFusion~\citep{Liang2022ECCV} & \underline{6.96}  & \underline{52.84}  & 11.49  & \underline{0.97} & 0.71 & 0.83  &  & 7.52  & 56.65  & 11.55  &  0.80 & 0.62  & 0.82  \\
			RFL~\citep{wang2022self} & 6.87  & 50.62  & 15.26  &  0.95 &  0.74 &  \underline{0.83} &   & 7.53  & 57.53  & 18.43  & \underline{0.94}  & \underline{0.68}  & 0.83  \\
			EPT~\citep{wang2023multi} & 6.87 &  50.64 & 15.30  & 0.95  & \underline{0.75}  & 0.82  &   &  \underline{7.53} & \underline{57.55}  & \underline{18.44}  & 0.94  & 0.68  &\textbf{0.85}  \\
			ReFusion (Ours) &  \textbf{6.98} & \textbf{54.35} &\textbf{15.65} & \textbf{1.09} & \textbf{0.76} & \textbf{0.83} &  & \textbf{7.56} & \textbf{59.82} &   \textbf{18.56} & \textbf{0.95} & \textbf{0.69} &  \underline{0.83} \\ \bottomrule
	\end{tabular}}
\end{table*}%

\subsection{Multi-focus Image Fusion}
For MFIF analysis, we use two primary datasets: RealMFF \citep{zhang2020real}, selecting 400 image pairs for training and 200 for testing, and Lytro~\citep{nejati2015multi}, with 20 image pairs for generalization assessment. Competitors include DIFNet~\citep{jung2020unsupervised}, CUNet~\citep{deng2020deep}, SDNet~\citep{DBLP:journals/ijcv/ZhangM21},  U2Fusion~\citep{9151265}, DeFusion~\citep{Liang2022ECCV}, RFL~\citep{wang2022self}, and EPT~\citep{wang2023multi}. The evaluation of MFIF performance is conducted using various metrics: entropy (EN), standard deviation (SD), spatial frequency (SF), visual information fidelity (VIF), human visual perception ($Q_{CB}$), and nonlinear correlation information entropy ($Q_{NCIE}$). These metrics collectively assess the quality and efficacy of fusion methods in preserving and enhancing image details across different focuses.
The qualitative analysis, as illustrated in 
Fig.~\ref{fig:MFIF}, highlights the ability of ReFusion to outperform other methods, showcasing its capability to not only preserve the rich textural details in the transitional areas between focused and unfocused regions but also maintain the clarity of pixels in both near-focus and far-focus images. Our method effectively selects and integrates clear regions from multiple-focus image pairs, ensuring that both foreground and background are presented clearly. This distinction is crucial, as reconstructing out-of-focus areas from a clear fused image is considerably easier, while a blurry composite significantly complicates detail recovery. This enhanced clarity is vital for applications requiring high precision in image analysis, such as surveillance and advanced photographic techniques.
The quantitative findings, detailed in Table~\ref{tab:MFIF}, confirm that ReFusion consistently delivers superior performance across various datasets. This achievement underscores the effectiveness of ReFusion in producing high-quality, clear images for MFIF, highlighting its capacity to integrate the best features from multiple sources into a single, coherent image.

\begin{figure*}[t]
	\centering
	\includegraphics[width=\linewidth]{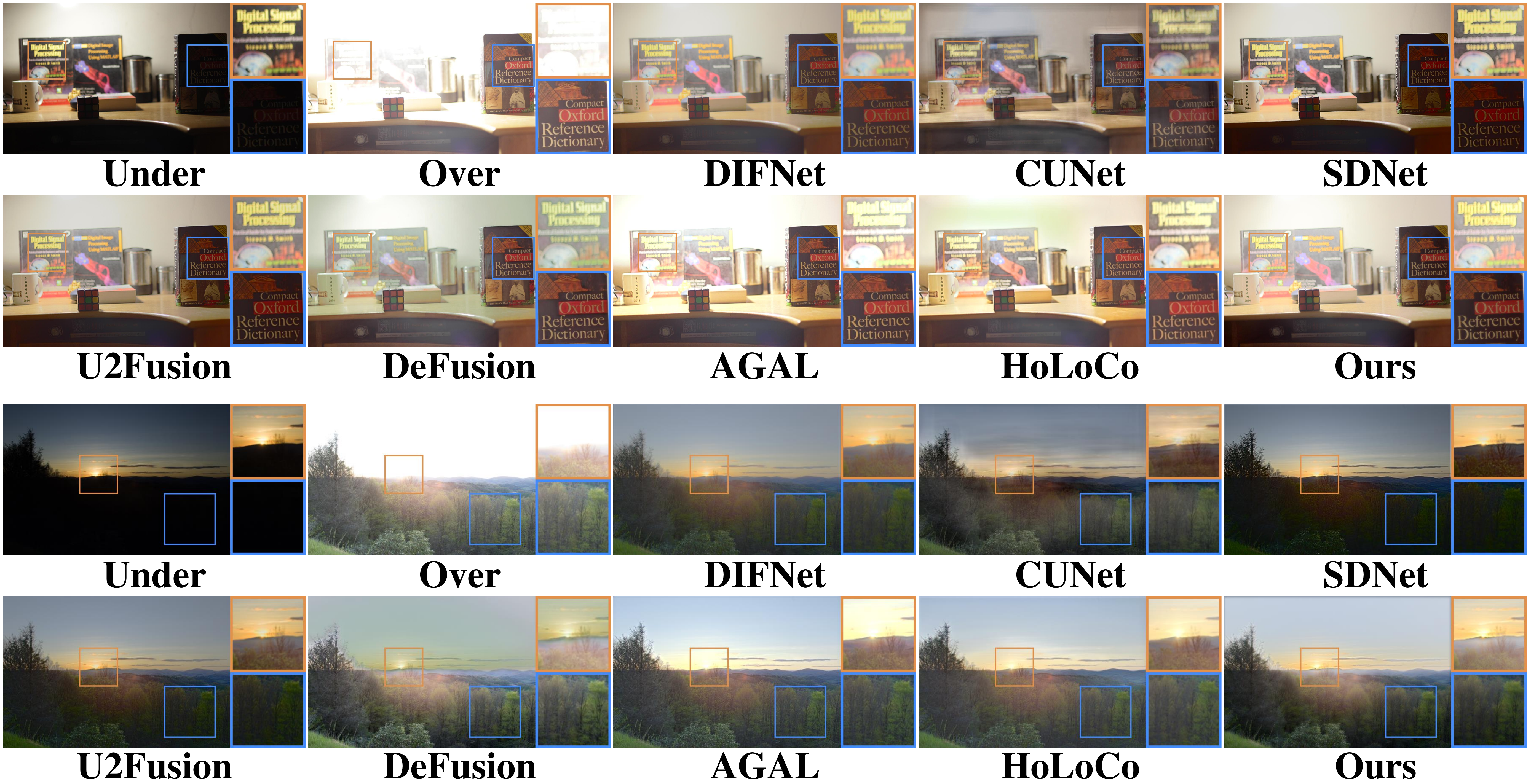}
	\caption{Visualization of the ``ICCV\_03'' and ``AirBellowsGap'' in MEFB dataset for MEIF.}
	\label{fig:MEIF}
\end{figure*}

\begin{table*}[t]
	\centering
	\caption{Quantitative comparison of MEIF. The best and second-best values are \textbf{highlighted} and \underline{underlined}.}
	\label{tab:MEIF}%
	\resizebox{\linewidth}{!}{
		\begin{tabular}{lccccccccccccc}
			\toprule
			
			&\multicolumn{6}{c}{\textbf{Multi-exposure Image Fusion on SICE Dataset}}                                      &   &                                          \multicolumn{6}{c}{\textbf{Multi-exposure Image Fusion on MEFB Dataset}}                                            \\
		Methods&  EN $\uparrow$   &   SD $\uparrow$   &  SF $\uparrow$   &  VIF $\uparrow$  & $Q_{CB}$ $\uparrow$  & $Q_{NCIE}$  $\uparrow$  &                                 &  EN $\uparrow$   &   SD $\uparrow$   &  SF $\uparrow$   &  VIF $\uparrow$  & $Q_{CB}$ $\uparrow$  & $Q_{NCIE}$  $\uparrow$  \\ \midrule
			DIFNet~\citep{jung2020unsupervised}  &  6.39 & 29.11  & 17.19  & 0.49  &  0.26 & \underline{0.82}  &    & 7.06 &  42.62 & 12.48  & 0.57  &  0.37 & \underline{0.83}  \\
			CUNet~\citep{deng2020deep}  & 6.87 & 32.28  & 16.22  & 0.79  & 0.37  & 0.81 &    & 7.14  & 38.84  & 12.75  & 0.71  & 0.41 &  0.81 \\
			SDNet~\citep{DBLP:journals/ijcv/ZhangM21}  & 6.61  & 34.94  & \textbf{29.41}  & 0.57  & \textbf{0.44}  &  0.81 &   & 7.01  & 48.12  & \textbf{23.60}  & 0.67  & \textbf{0.49}  & 0.82  \\
			U2Fusion~\citep{9151265} & 6.38  & 29.30  & 14.23 & 0.53  & 0.36  & 0.82  &   & 6.89  & 41.22  & 13.27 & 0.60  & 0.46  & 0.83  \\
			DeFusion~\citep{Liang2022ECCV} & 6.61  & 36.60  & 16.55  & 0.77  & 0.28  & 0.81 &   & 6.91  & 45.70  & 13.71  & 0.67  & 0.37  & 0.82  \\
			AGAL~\citep{liu2022attention} & \underline{7.02}  & \underline{39.70}  & 23.60  & \underline{0.80}  & 0.38  & 0.81  &   &  \underline{7.30} & \underline{52.02}  & 18.53  &  \underline{0.83} & 0.45  & 0.83  \\
			HoLoCo~\citep{liu2023holoco} & 6.98  &  37.07 & 11.69  &  0.80 & 0.40  & 0.81  &   & 7.24 & 46.10  & 13.26  & 0.76  & 0.45  & 0.82  \\
			ReFusion (Ours) & \textbf{7.38} & \textbf{55.48} &  \underline{23.69} & \textbf{1.18} & \underline{0.41} & \textbf{0.82}  &   & \textbf{7.47} & \textbf{62.83} &  \underline{19.01} & \textbf{0.97} & \underline{0.47} &\textbf{0.83} \\ \bottomrule
	\end{tabular}}
 \vspace{-1em}
\end{table*}%

\subsection{Multi-exposure Image Fusion}
In our MEIF study, we utilize a training set comprising 300 images randomly chosen from SICE~\citep{cai2018learning}, and a combined test set of 100 images from SICE~\citep{cai2018learning} and 30 from MEFB~\citep{zhang2021benchmarking} for evaluation. The competing methodologies include DIFNet~\citep{jung2020unsupervised}, CUNet~\citep{deng2020deep}, SDNet~\citep{DBLP:journals/ijcv/ZhangM21}, U2Fusion~\citep{9151265}, DeFusion~\citep{Liang2022ECCV}, AGAL~\citep{liu2022attention}, HoLoCo~\citep{liu2023holoco}. Given the shared domain of digital image fusion, we employ the same evaluation metrics for MEIF as those used in MFIF, ensuring consistency in performance assessment across different image fusion tasks.
The visualization outcomes, as depicted in Fig.~\ref{fig:MEIF}, illustrate the significant advantages of ReFusion.
ReFusion notably enhances the brightness and contrast in areas of extreme exposure and amplifies color saturation, while simultaneously minimizing edge blurring artifacts. A critical advantage of our method over competitors is its ability to avoid inaccurate color restoration. These enhancements contribute to producing images that are visually closer to natural human perception, underscoring the effectiveness of ReFusion in handling complex exposure scenarios. This capability makes it particularly valuable in applications such as digital photography enhancement, video production, and any field requiring high-quality visual representations under varying lighting conditions.
The quantitative analysis in Table~\ref{tab:MEIF} shows that ReFusion delivers high-quality outcomes with precise texture and intensity distributions. This substantiates ReFusion's capacity to produce images that accurately represent the original scenes, affirming its superiority in the MEIF task. This capability ensures that fused images retain critical visual information, enhancing accuracy and quality in scenarios with significant exposure variations, such as in photography and surveillance applications.

\begin{figure*}[t]
	\centering
	\includegraphics[width=\linewidth]{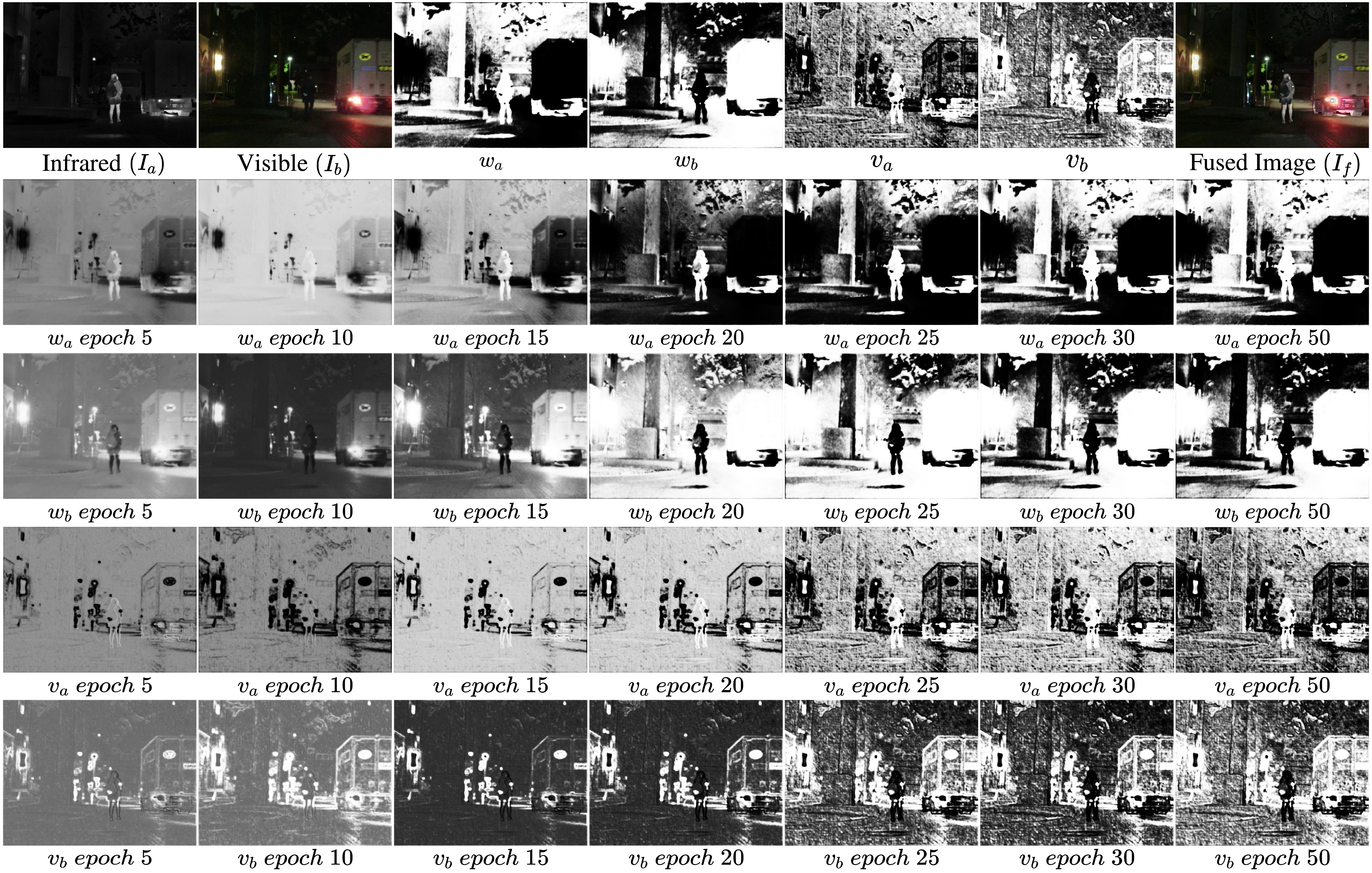}
	\caption{Learnable fusion loss of ``00714N'' in MSRS dataset for infrared-visible image fusion. Both the intensity part $w$ and the gradient part $v$ of the fusion loss tend to favor the parts of the source image that have more information. For example, the human part of the infrared image has a larger $w_a$ and $v_a$, while the building on the left of the image and the truck on the right have a larger $w_b$ and $v_b$.}
	\label{fig:LL_MSRS_00714N}
\end{figure*}
\begin{figure*}[t]
	\centering
	\includegraphics[width=\linewidth]{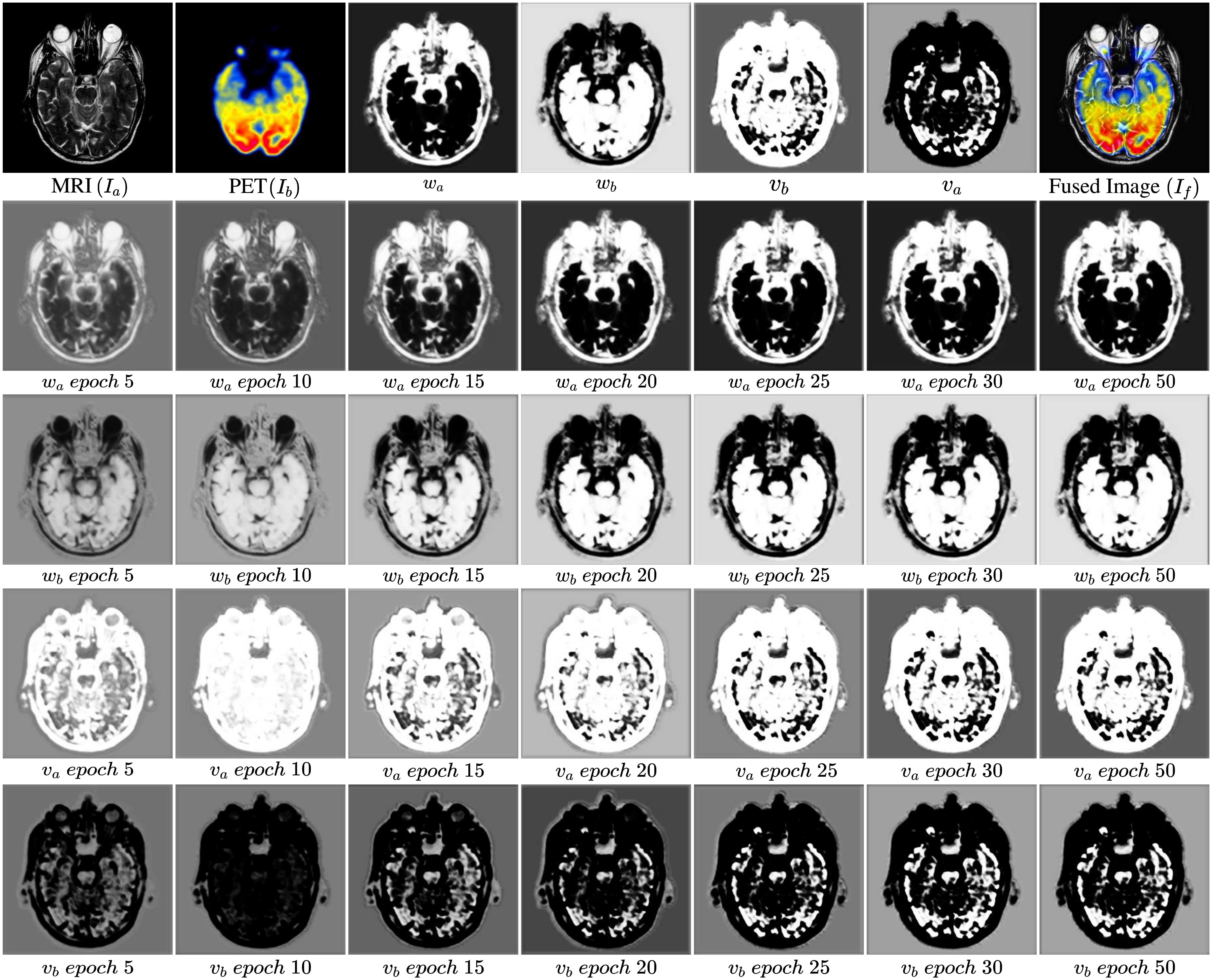}
	\caption{Learnable loss of MRI-PET case in Harvard dataset for medical image fusion. Fusion loss attempts to preserve both the histological information in MRI image and the functional information in PET image.}
	\label{fig:LL_MIF_MRI_PET_37}
	\end{figure*}

	\begin{figure*}[t]
        
	\centering
	\includegraphics[width=\linewidth]{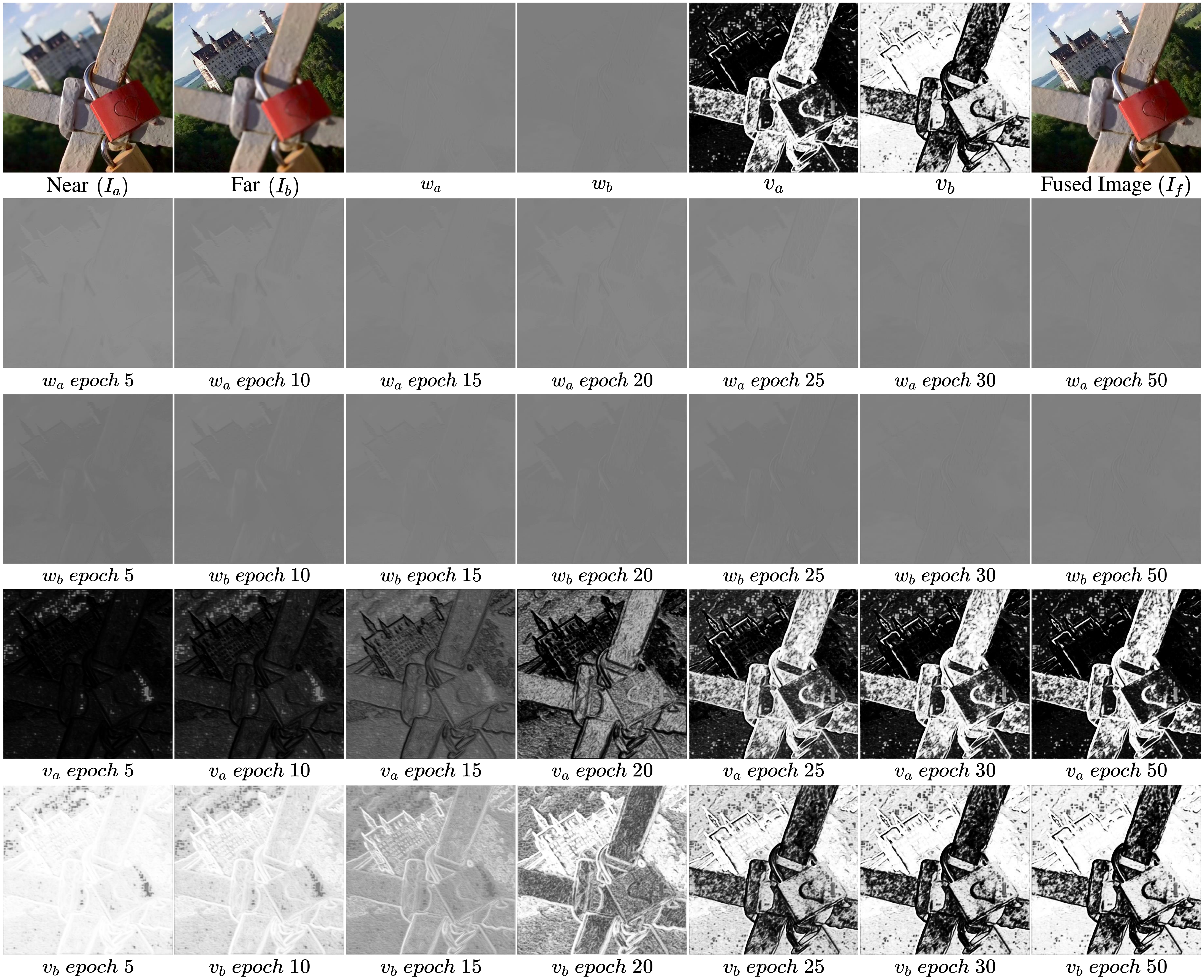}
	\caption{Learnable loss of ``Lytro\_6'' in Lytro dataset for multi-focus image fusion. The intensity of the two source images is extremely similar, so the intensity part of the fusion loss has almost the same $w$. The difference in the clarity of the two source images results in a difference in the gradient part of the fusion loss, with the sharper portion of the source image having a larger gradient weight $v$.}
	\label{fig:LL_Lytro_6}
        \vspace{-0.5em}
	\end{figure*}

	\begin{figure*}[t]
	\centering
	\includegraphics[width=\linewidth]{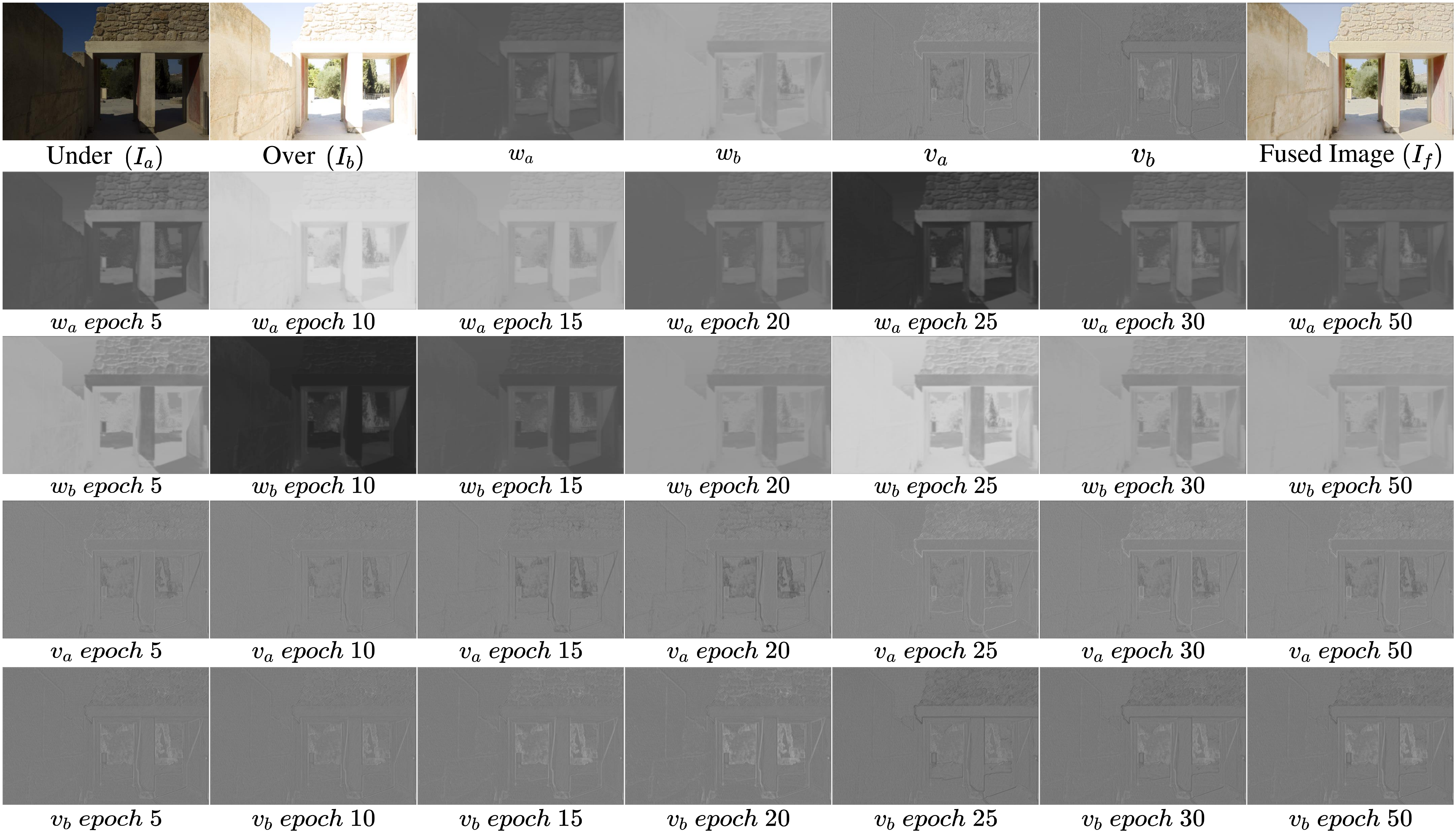}
	\caption{Learnable loss of ``Knossos7'' in MEFB dataset for multi-exposure image fusion. The adaptive results for fusion loss show that in the multi-exposure image fusion task. A near-equal degree of intensity preference keeps the fusion image at an intensity level that maximizes the retention of information from the source image. And for the lost gradient information due to overexposure or underexposure, the gradient part $v$ of the fusion loss preserves the information from the other image accordingly.}
	\label{fig:LL_MEFB_Knossos7}
	\end{figure*}
    
\subsection{Learnable Loss} 
Our approach introduces a novel strategy for image fusion by parameterizing the fusion loss and utilizing a loss proposal module that predicts the optimal preferences from the source images.
The learnable fusion loss parameters, $w$ and $v$, adaptively adjust to meet the needs of specific fusion tasks. The dynamic adjustment process and learning results are displayed in Figs.~\ref{fig:LL_MSRS_00714N}, \ref{fig:LL_MIF_MRI_PET_37}, \ref{fig:LL_Lytro_6}, \ref{fig:LL_MEFB_Knossos7}.

\begin{figure*}[h]
	\centering
	\includegraphics[width=\linewidth]{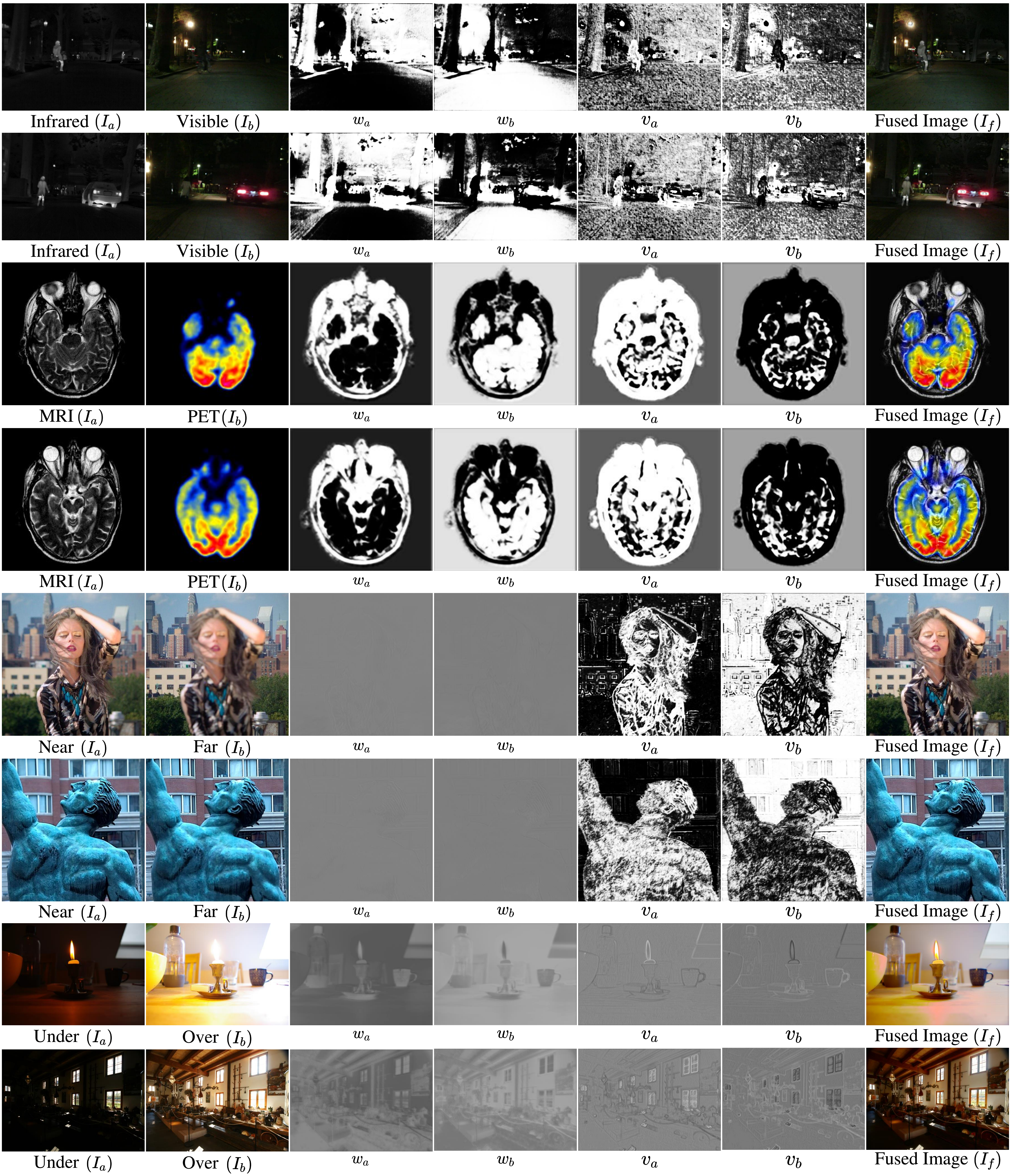}
	
	\caption{{More visualization examples of learnable fusion loss across different fusion tasks.}}
	\label{fig:more_LL}

    \clearpage{}
\end{figure*}

For IVIF, as shown in Fig.~\ref{fig:LL_MSRS_00714N}, the learnable loss is performed to capture thermal radiation details from infrared images while retaining critical visual information from visible images. This ensures a comprehensive fusion that highlights both thermal properties and visible features.
In MIF, as shown in Fig.~\ref{fig:LL_MIF_MRI_PET_37}, the learning loss adapts to preserve essential histological and functional information across the images, selectively integrating the most relevant details from each source image into a detailed composite image.
For MFIF, as shown in Fig.~\ref{fig:LL_Lytro_6}, the focus of fusion loss shifts towards the gradient differences between source images, where $v$ indicates a preference for preserving clarity through gradient prioritization, adjusting based on the variance in focus clarity across the images.
In MEIF, as shown in Fig.~\ref{fig:LL_MEFB_Knossos7}, the learnable loss concentrates on identifying and weighting the most informative parts of each source image. This process ensures that the intensity levels of fused image are optimally adjusted to encapsulate the maximum amount of information possible from the source images.
In addition to understanding the learning process of learnable loss, examples of additional learning outcomes are displayed in Fig.~\ref{fig:more_LL}, to better comprehend the functionality and efficacy of the learnable loss proposed in this study.

\begin{table*}[h]
	\centering
	\centering
	\caption{Ablation experiment results of IVIF. \textbf{Bold} indicates the best value.}
	\label{tab:Ablation}
	\resizebox{0.8\linewidth}{!}{
		\begin{tabular}{cccccccc}
			\toprule
			\multicolumn{8}{c}{\textbf{Ablation Studies of IVIF on MSRS Dataset}}  
			\\
			&                       {Configurations}                       &        EN        &        SD          &        SF               &       AG       &       SCD        &    SSIM   \\ \midrule
			\uppercase\expandafter{\romannumeral1} & w/o Learnable loss, fix $w$ and $v$ as $1/2$ &     6.26      &     37.49      &     10.86      &     2.69   &1.42&  0.37 \\
			\uppercase\expandafter{\romannumeral2} &          w/o Training LPM based on meta-learning             &     6.23      &     38.96     &     10.43     &     2.15  &1.37&  0.47  \\
			\uppercase\expandafter{\romannumeral3} &              w/o Fusion and Reconstruction update              &     6.25      &     39.33     &     10.71     &     2.99  &1.33&  0.46  \\ 
			\uppercase\expandafter{\romannumeral4} &             max $\rightarrow$ sum in Eq.~(\ref{eq5}) and Eq.~(\ref{eq6})                  &     6.03      &     32.65      &     10.75      &    2.49   &1.27& 0.33 \\
			
			\uppercase\expandafter{\romannumeral5} &              $I_f=w_a*I_a+w_b*I_b$                  &     6.32      &     38.42      &     11.03      &    2.89   &1.43& 0.42 \\
			\uppercase\expandafter{\romannumeral6} &               AFM $\rightarrow$ Concatenate               &     6.47      &     39.26      &     10.93          &2.94&     1.46& 0.39 \\
			\uppercase\expandafter{\romannumeral7} &              w/o Interactive feature extraction              &     6.52      &     39.25     &     10.96       &3.01&     1.54&  0.49  \\
			\uppercase\expandafter{\romannumeral8} &              w/o  Gating Feature Refinement              &     6.55      &     39.53      &     10.93           &2.98&     1.62&0.48 \\  \midrule

			&                            {Ours}                           &  \textbf{6.71} & \textbf{42.89} & \textbf{11.64} & \textbf{3.80} & \textbf{1.67} & \textbf{0.51} \\ \bottomrule
	\end{tabular}}
\end{table*}

\begin{figure*}[t]
	\centering
	\includegraphics[width=0.95\linewidth]{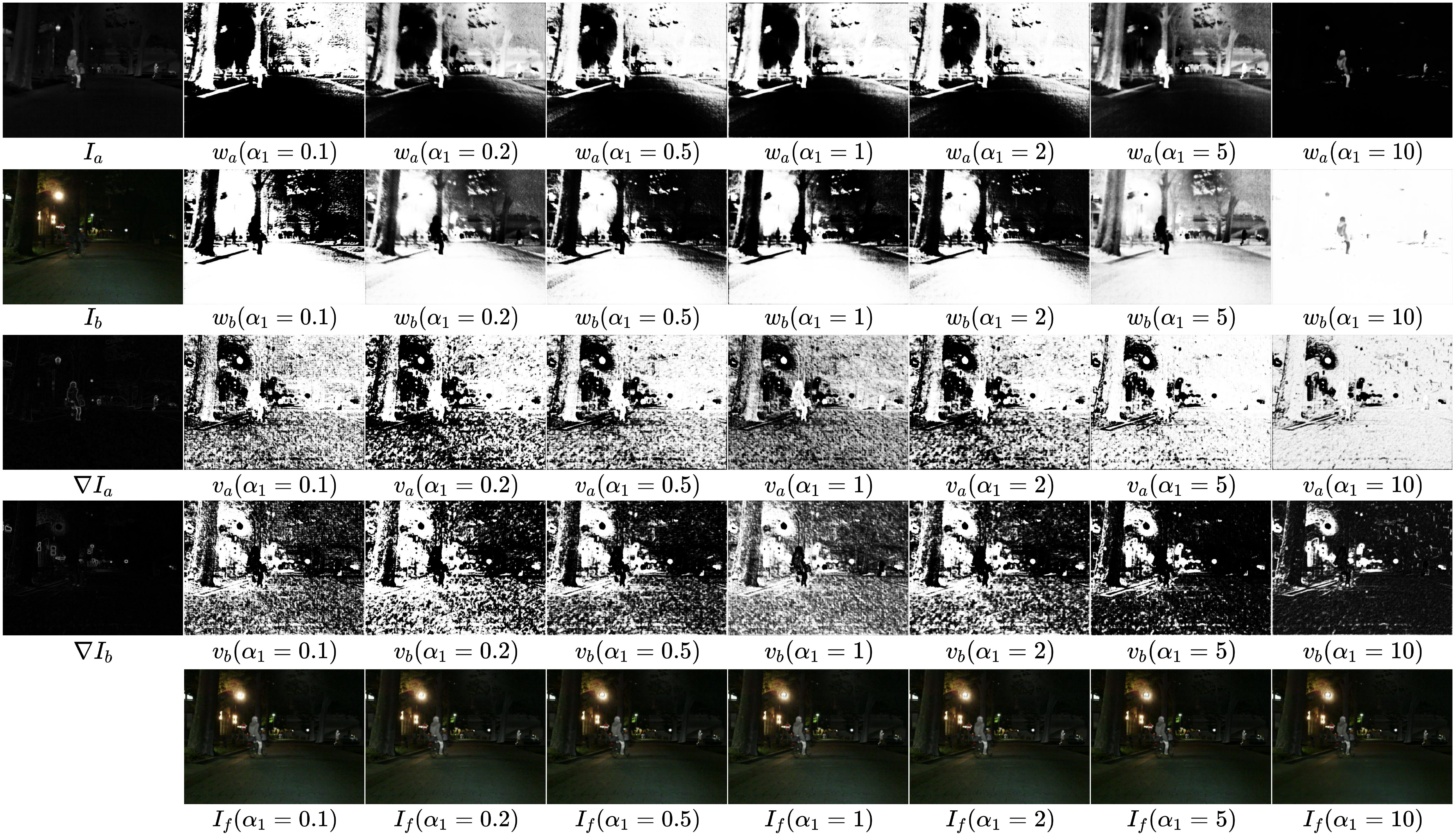}

	\caption{{Hyperparameter selection for $\alpha_1$ with $\alpha_2=1$, visualization of fused image $I_f$ and learnable fusion loss $\{w_a,w_b,v_a,v_b\}$ for different $\alpha_1$}.}
	\label{fig:alpha1}
\end{figure*}
\begin{figure*}[t]
	\centering
	\includegraphics[width=0.95\linewidth]{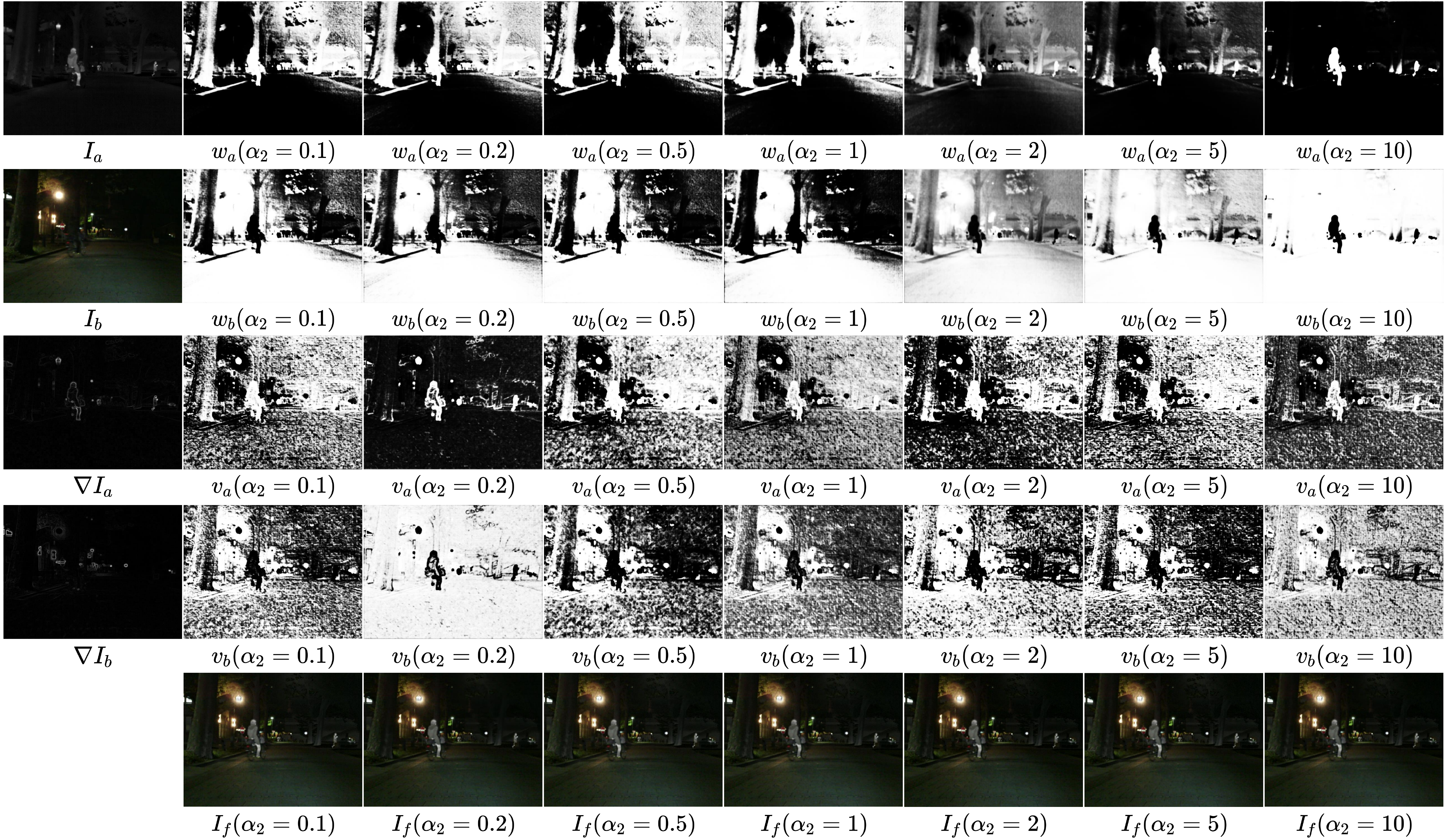}

	\caption{{Hyperparameter selection for $\alpha_2$ with $\alpha_1=1$, visualization of fused image $I_f$ and learnable fusion loss $\{w_a,w_b,v_a,v_b\}$ for different $\alpha_2$}.}
	\label{fig:alpha2}
\end{figure*}

\subsection{Ablation Studies} 
We conduct ablation experiments using IVIF as an example to evaluate the impact of our training strategy and the structural components of our method. The results can be found in Table~\ref{tab:Ablation}.
In Exp.~\uppercase\expandafter{\romannumeral1}, fixing the learnable loss weights $w$ and $v$ to $1/2$ limits the loss function's adaptability, thereby reducing performance.
In Exp.~\uppercase\expandafter{\romannumeral2}, omitting the meta-learning strategy and simultaneously employing both the $\mathcal{L}_f$ in Exp.~\uppercase\expandafter{\romannumeral1} and $\mathcal{L}_r$ for training all modules reveal that this direct approach yields suboptimal results, indicating the importance of a more nuanced training methodology.
In Exp.~\uppercase\expandafter{\romannumeral3}, allowing inner updates to influence $\mathcal{F}$ without the updates from fusion and reconstruction reduces system effectiveness, underscoring the importance of these updates in our strategy.
In Exp.~\uppercase\expandafter{\romannumeral4}, switching the $Max(\cdot)$ operation in the $\mathcal{L}_r$ with $Sum(\cdot)$ leads to a degradation in performance. This change makes $\mathcal{L}_f$ less effective at integrating information from source images, veering $\mathcal{F}$ towards prioritizing the image with the better initial reconstruction.
In Exp.~\uppercase\expandafter{\romannumeral5}, transitioning to a decision-based fusion approach while training in a manner akin to ReFusion doesn't produce the desired high-quality images, highlighting the inadequacy of a purely decision-based method in this context.
In Exp.~\uppercase\expandafter{\romannumeral6}-Exp.~\uppercase\expandafter{\romannumeral8},  we made several modifications to the structure of $\mathcal{F}$, including replacing the Attention Fusion Module (AFM) with simple concatenation, removing the interactive extraction module, and excluding the gating refinement module. All these changes led to diminished results. These changes confirm the integral role of these components in achieving optimal fusion outcomes.
Through these ablation experiments, we conclusively demonstrate the effectiveness and necessity of the chosen training strategies and structural elements.

\subsection{Hyperparameter selection} 
\begin{table*}[t]
	\caption{{Hyperparameter selection for $\alpha_1$ and $\alpha_2$ on MSRS dataset. The best and second-best values are \textbf{highlighted} and \underline{underlined}.}}
	\label{tab:alpha}
	\centering
	\resizebox{0.9\linewidth}{!}{
		\begin{tabular}{cccccccccccccccc}
			\toprule
			\multicolumn{7}{c}{\textbf{Hyperparameter selection for $\alpha_1$ $(\alpha_2=1)$}}&&&
			\multicolumn{7}{c}{\textbf{Hyperparameter selection for $\alpha_2$ $(\alpha_1=1)$}}\\
			$\alpha_1$ & EN   & SD    & SF    & AG   & SCD  & SSIM &&& $\alpha_2$ & EN   & SD    & SF    & AG   & SCD  & SSIM \\ \midrule
			0.1    & 6.67 & 42.67 & 11.58 & 3.68 & 1.67 & 0.49 &&& 0.1    & \underline{6.71} & 42.72 & 11.63 & 3.78 & 1.66 & 0.51 \\ 
			0.2    & 6.69 & 42.72 & 11.58 & 3.71 & 1.66 & 0.48 &&& 0.2    &  6.69& 42.67 & 11.47 & 3.72 & 1.66 & 0.49 \\ 
			0.5    & \underline{6.71} & 42.72 & 11.61 & 3.76 & \textbf{1.68} & \underline{0.51} &&& 0.5    & {6.70}  & 42.85 & 11.56 & 3.75 & \textbf{1.68} & \underline{0.51} \\ 
			1      & \textbf{6.71} & \underline{42.89} & 11.64 & 3.80 & \underline{1.67} & \textbf{0.51} &&& 1      & \textbf{6.71} & \textbf{42.89} & \underline{11.64} & \underline{3.80} & \underline{1.67} & \textbf{0.51} \\ 
			2      & 6.70  & \textbf{43.16} & \textbf{11.67} & 3.81 & 1.66 & 0.49 &&& 2      & 6.69 & \underline{42.89} & \textbf{11.66} & \textbf{3.83} & 1.66 & 0.50  \\ 
			5      & 6.70  & 42.77 & 11.65 & \textbf{3.82} & 1.64 & 0.48 &&& 5      & 6.68 & 42.88 & 11.59 & 3.78 & 1.66 & 0.49 \\ 
			10     & 6.65 & 41.04 & \underline{11.67} & \underline{3.82} & 1.61 & 0.45 &&& 10     & 6.68 & 42.78 & 11.64 & 3.79 & 1.64 & 0.49 \\ \bottomrule
	\end{tabular}}
\vspace{-1em}
\end{table*}

{Our framework includes two hyperparameters, $\alpha_1$ and $\alpha_2$, used to balance the intensity and gradient terms in the fusion loss and reconstruction loss, respectively. We explored the effects of these hyperparameters on fusion performance through experiments with IVIF as the baseline. The comparative training results for $\alpha_1$ and $\alpha_2$ are displayed in Fig.~\ref{fig:alpha1} and Fig.~\ref{fig:alpha2}, respectively. The quantitative outcomes of these experiments are detailed in Table~\ref{tab:alpha}.}

{The results from Table~\ref{tab:alpha} show that increasing $\alpha_1$ affects the learning of the intensity term in the fusion loss, leading to fused images with lower contrast in their intensity components. This is quantitatively manifested as a decline in the Entropy (EN) and Standard Deviation (SD). On the other hand, a very low $\alpha_1$ diminishes the texture retention ability of the fusion network. A noticeable impact of this is the disappearance of street lamp contours in the images as $\alpha_1$ decreases. Therefore, to achieve optimal performance, $\alpha_1$ should ideally be set between 0.5 and 2. In our experiments, we set $\alpha_1$ at 1.}

{The influence of hyperparameter $\alpha_2$ in the reconstruction loss is relatively minor compared to $\alpha_1$ in the fusion loss. Visual comparisons in Fig.~\ref{fig:alpha2} demonstrate that under various conditions of $\alpha_2$, nearly indistinguishable fusion results are generated to the naked eye, with only a slight impact on the learning of intensity loss weight when $\alpha_2$ increases to 10. The research presented in Table~\ref{tab:alpha} also confirms that various settings of $\alpha_2$ yield similar fusion performances. The optimal range for $\alpha_2$ is from 0.2 to 5. In our experiments, $\alpha_2$ is set to 1.}

\begin{table*}[t]
	\centering
	\caption{Trainable parameter comparison of the competitors. The \colorbox{firstcolor}{red} markers represent the methods with fewer trainable parameters than ReFusion.}
	\label{tab:para}%
	\resizebox{0.8\linewidth}{!}{
		\begin{tabular}{lclclclc}
			
			\toprule
			\multicolumn{8}{c}{\textbf{Parameter comparison for Fusion Modules} (\textit{M})}	\\
			Methods &Para& 
			Methods &Para& 
			Methods &Para& 
			Methods &Para
			
			\\ \midrule
			DIF-Net~\citep{jung2020unsupervised} & \colorbox{firstcolor}{0.022} &
			RFL~\citep{wang2022self}  & 15.859&
			CDDFuse~\citep{zhao2023cddfuse} &1.188&
			GeSeNet~\citep{li2023gesenet}&0.241 \\
			
			CUNet~\citep{deng2020deep}   &0.638&
			AGAL~\citep{liu2022attention}   &1.591 &
			LRRNet~\citep{li2023lrrnet}&\colorbox{firstcolor}{0.049}&
			MsgFusion~\citep{wen2023msgfusion}&2.971\\
			
			SDNet~\citep{DBLP:journals/ijcv/ZhangM21}  &0.067&
			TarDAL~\citep{DBLP:conf/cvpr/LiuFHWLZL22}   &0.297&
			MURF~\citep{xu2023murf}&0.27&
			EPT~\citep{wang2023multi}&  1.897\\
			
			U2Fusion~\citep{9151265}   &0.659&
			DeFusion~\citep{Liang2022ECCV}   &7.874&
			DDFM~\citep{zhao2023ddfm}&/&
			HoLoCo~\citep{liu2023holoco}&17.691 \\
			
			MATR~\citep{tang2022matr} & \colorbox{firstcolor}{0.013}&
			MetaFusion~\citep{zhao2023metafusion}   &0.812&
			SegMIF~\citep{Liu_2023_ICCV}&1.034&
			\textbf{ReFusion (Ours)}& \textbf{0.060} \\
			\bottomrule
	\end{tabular}}%
\end{table*}

\subsection{Comparison of trainable parameters}
The comparison of the trainable parameters across the methods discussed in this paper is detailed in Table~\ref{tab:para}. The red markers highlight the methods with fewer parameters than ReFusion. Among these, DIFNet~\citep{jung2020unsupervised}, MATR~\citep{tang2022matr}, and LRRNet~\citep{li2023lrrnet} are the only methods with fewer parameters than ReFusion, with the exception of DDFM~\citep{zhao2023ddfm}, which does not require training.
This distinction underscores the efficiency of ReFusion, combining a lightweight design with robust fusion capabilities. 
Our method's architecture ensures high-quality fusion outcomes and meets real-time processing demands, making it ideal for applications that require immediate computational feedback.

\section{Conclusion}
We propose a unified image fusion framework based on meta-learning, where the algorithm acquires the fusion capability from the source images reconstruction. This approach addresses the inherent challenge of image fusion tasks, which lack ground truth and require manually designed loss functions.
In this paper, reconstruction loss of the source images is used as a criterion for information preservation, aiming to retain as much information from the source images as possible in the fused image.
The loss proposal module is then trained via meta-learning using source reconstruction loss. 
This module can propose suitable fusion loss for different tasks, thereby more effectively guiding the fusion network training.
The meticulously designed lightweight fusion module, coupled with adaptive fusion loss, successfully completes a variety of fusion tasks with high quality, including IVIF, MIF, MFIF and MEIF. 
Theoretical analysis and experimental results demonstrate the rationality, superiority and versatility of our framework.

{
    \small
    \bibliographystyle{ieeenat_fullname}
    \bibliography{refer}
}


\end{document}